\pdfoutput=1

\documentclass[11pt,table,dvipsnames]{article}

\usepackage[final]{acl}

\usepackage{times}
\usepackage{latexsym}

\usepackage[T1]{fontenc}

\usepackage[utf8]{inputenc}

\usepackage{microtype}

\usepackage{inconsolata}

\usepackage{graphicx}
\usepackage{colortbl}
\usepackage{color,xcolor}
\usepackage{graphicx}
\usepackage{inconsolata}
\usepackage{microtype}
\usepackage{latexsym}
\usepackage{dsfont}
\usepackage{booktabs} 
\usepackage{CJKutf8}
\usepackage{graphicx}
\usepackage{bigstrut}
\usepackage{multirow}
\usepackage{amsmath}
\usepackage{multirow, makecell, caption}
\usepackage{floatrow}
\newfloatcommand{capbtabbox}{table}[][\FBwidth]
\usepackage{xcolor}
\usepackage[normalem]{ulem}
\usepackage{amssymb}
\usepackage{amsfonts}
\usepackage{multicol}
\usepackage{tipa}
\usepackage{amsmath}
\usepackage{bm}
\usepackage[ruled,linesnumbered]{algorithm2e}
\usepackage{tikz}
\usepackage{paralist}
\usepackage{IEEEtrantools}
\usepackage[switch]{lineno}
\usepackage{enumitem}
\usepackage{multirow, makecell, caption}
\usepackage{arydshln}
\usepackage{verbatim}
\usepackage[normalem]{ulem}
\usepackage{amssymb}
\usepackage{amsfonts}
\usepackage{multicol}
\usepackage{amssymb}
\usepackage{pifont}
\usepackage{csquotes}
\usepackage{xspace}
\usepackage{paralist}
\usepackage{mdwlist}
\usepackage{subcaption}
\usepackage{makecell}
\usepackage{array}
\usepackage{bm}
\usepackage{colortbl}
\usepackage{dsfont}
\usepackage{booktabs} 
\usepackage{graphicx}
\usepackage{tabularx}
\usepackage{amsmath}
\usepackage{bbm}
\usepackage{pgfplots}
\usepackage{soul} 
\usepackage{listings}
\usepackage[tableposition=top]{caption}
\usepackage{color,xcolor} 
\usepackage{footmisc}
\usepackage{xurl}

\definecolor{color1}{RGB}{210, 235, 180}
\definecolor{color2}{RGB}{240, 190, 140}
\definecolor{color3}{RGB}{220, 200, 230}
\definecolor{color4}{RGB}{150, 170, 210}

\newcommand{\ie}{\textit{i.e.}\xspace}
\newcommand{\eg}{\textit{e.g.}\xspace}

\newcommand{\method}{\textsc{HisAnaloy}\xspace}\makeatletter
\def\adl@drawiv#1#2#3{%
        \hskip.5\tabcolsep
        \xleaders#3{#2.5\@tempdimb #1{1}#2.5\@tempdimb}%
                #2\z@ plus1fil minus1fil\relax
        \hskip.5\tabcolsep}
\newcommand{\cdashlinelr}[1]{%
  \noalign{\vskip\aboverulesep
           \global\let\@dashdrawstore\adl@draw
           \global\let\adl@draw\adl@drawiv}
  \cdashline{#1}
  \noalign{\global\let\adl@draw\@dashdrawstore
           \vskip\belowrulesep}}
\makeatother

\title{\textit{Past Meets Present}:  \\ Creating Historical Analogy with Large Language Models}

\author{Nianqi Li$^{1}$, Siyu Yuan$^{2}$\thanks{Corresponding author.}, Jiangjie Chen$^{3}$\thanks{Part of the work done while at Fudan Univeristy.}, \\ \bf Jiaqing Liang$^{2}$, Feng Wei$^{4}$, Zujie Liang$^{4}$, Deqing Yang$^{2}$, Yanghua Xiao$^{1}$\footnotemark[1] \\
$^{1}$Shanghai Key Laboratory of Data Science,\\ College of Computer Science and Artificial Intelligence, Fudan University\\
$^{2}$School of Data Science, Fudan University $^{3}$ByteDance Seed $^{4}$MYbank, Ant Group\\
\texttt{\{nqli23,syyuan21\}@m.fudan.edu.cn, shawyh@fudan.edu.cn}
}

\begin{document}
\maketitle
\begin{abstract}
Historical analogies, which compare known past events with contemporary but unfamiliar events, are important abilities that help people make decisions and understand the world.
However, research in applied history suggests that people have difficulty finding appropriate analogies. 
And previous studies in the AI community have also overlooked historical analogies.
To fill this gap, in this paper, we focus on the \textbf{historical analogy acquisition} task, which aims to acquire analogous historical events for a given event.
We explore retrieval and generation methods for acquiring historical analogies based on different large language models (LLMs).
Furthermore, we propose a self-reflection method to mitigate hallucinations and stereotypes when LLMs generate historical analogies.
Through human evaluations and our specially designed automatic multi-dimensional assessment, we find that LLMs generally have a good potential for historical analogies. 
And the performance of the models can be further improved by using our self-reflection method.\footnote{Resources of this paper can be found at \url{https://github.com/Nianqi-Li/Historical-Analogy-of-LLMs}}
\end{abstract}

\section{Introduction}
\label{sec:intro}
\begin{figure}[t]
    \centering
    \includegraphics[width=1.0\linewidth]{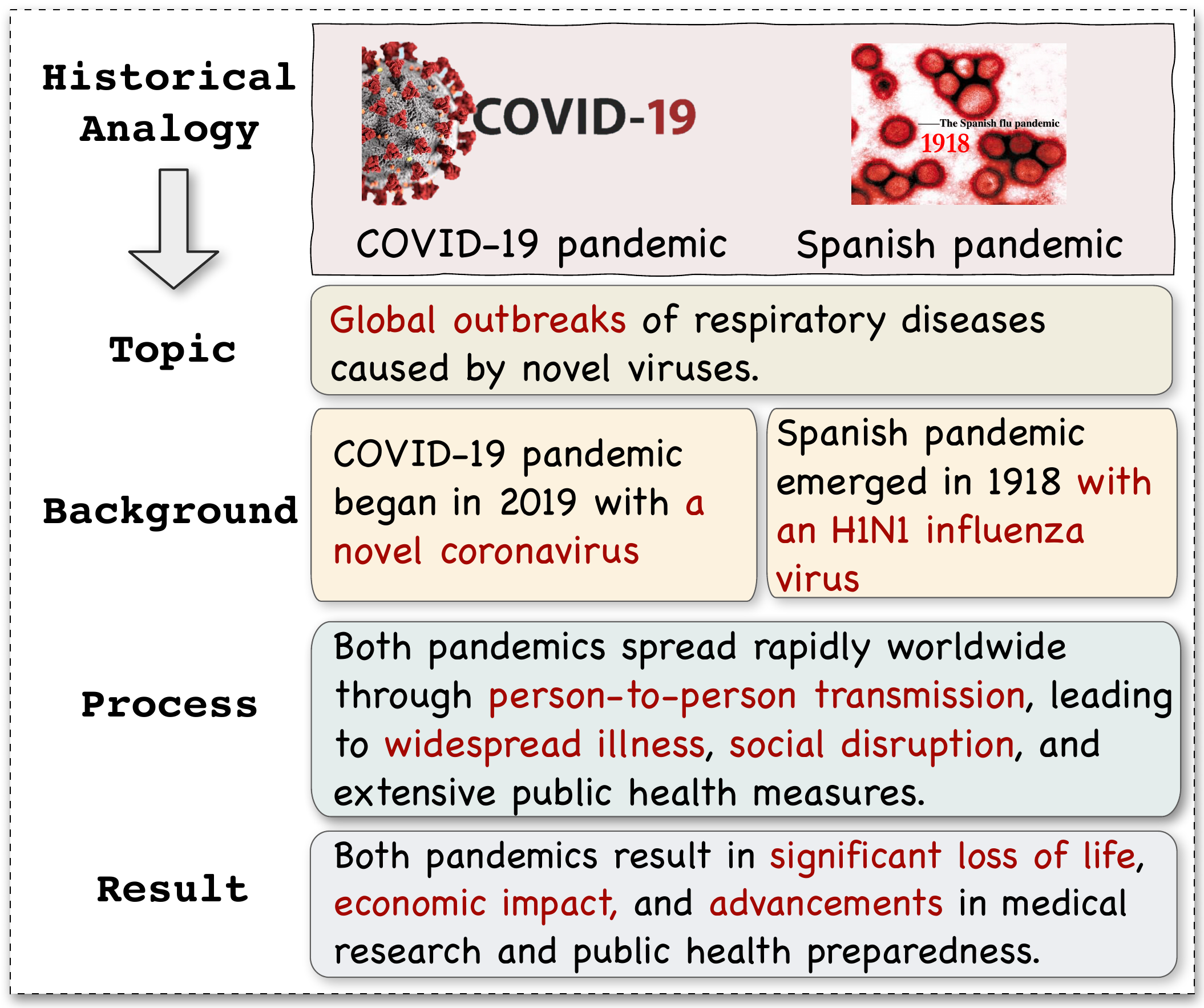}
    \caption{
    An example of historical analogy: analogizing COVID-19 to the Spanish pandemic based on topic, background, process, and result.
    }
    \label{fig:front}
\end{figure}

Historical analogy, which draws comparisons between contemporary and past situations, is a vital tool in applied history~\cite{achenbaum1983making,guldi2014history,parsons2016historical,ghilani2017looking,HistoricalAnalogies}. 
These analogies enable a deeper understanding of historical events and facilitate informed decision-making in addressing present difficulties~\cite{bartha2013analogy,AXELROD20178}.
For example, as shown in Figure~\ref{fig:front}, when the COVID-19 pandemic spread around the world, the influenza pandemic of 1918 emerged as an analogy, aiding in the navigation of the crisis.
However, historians have found that individuals, particularly politicians, often misuse historical analogies. 
They tend to gravitate towards the first analogy that comes to mind, are influenced by superficial similarities, and rarely conduct thorough analyses~\cite{ghilani2017looking,khong2020analogies}.
Furthermore, the creation of historical analogies involves having extensive knowledge of historical events and selecting the appropriate one, which can also be a great challenge.
Therefore, exploring large language models (LLMs)~\cite{llama3modelcard,openai2022chatgpt,openai2023gpt4} with the ability to automatically generate historical analogies is of great value.

Traditional studies within the AI community have concentrated on recognizing and generating word analogies, \eg, ``king is to man as queen is to woman'', using word embeddings~\cite{mikolov2013efficient,gladkova-etal-2016-analogy,fournier-etal-2020-analogies,ushio-etal-2021-bert} or by training language models (LMs)~\cite{czinczoll-etal-2022-scientific,chen-etal-2022-e,yuan-etal-2023-beneath}. 
Recently, with the advancement of LLMs, some researchers have designed prompts to instruct LLMs to generate free-form analogies~\cite{webb2022emergent}.  
However, these efforts are limited to the scientific domain (\eg, analogies for atom structure)~\cite{bhavya-etal-2022-analogy,sultan-shahaf-2022-life,jiayang-etal-2023-storyanalogy,yuan-etal-2023-beneath,sultan2024parallelparc,yuan2024boosting} or to everyday scenarios derived from webpages~\cite{wijesiriwardene-etal-2023-analogical,ding2023fluid,bhavya-etal-2024-anade1}, neglecting the exploration of analogies that draw comparisons between contemporary and historical situations, which could provide a comprehensible perspective on history.

In this paper, we explore the concept of historical analogy and introduce a new task, \ie, \textit{historical analogy acquisition}. 
This task aims to find historical events analogous to current events.
Specifically, given an event's name and text description, the ultimate goal is to obtain another historical event analogous to the original event in multiple dimensions, such as cause, process, and result.
To test the performance of LLMs on this task, we employ various methods based on two paradigms: 1) dataset retrieval methods, which employ LLMs to retrieve historical events from a specified dataset, and 2) free generation methods, which instruct LLMs to autonomously generate analogous historical events, leveraging the knowledge stored in their parameters. 
Furthermore, to mitigate the hallucination and stereotyping in generating historical analogies, we propose the self-reflection method, which comprises two LLM-based modules: the Candidate Generator and the Answer Reflector.
The Candidate Generator produces potential analogies, while the Answer Reflector offers feedback to refine these candidates to get rid of stereotypes.
Additionally, we verify the candidates through Wikipedia API to ensure their authenticity.

For evaluation, we employ both human and automatic methods to thoroughly assess the quality of historical analogies. 
In human evaluation, we use a manual ranking system to examine historical analogies. 
To reduce labor, we also introduce automatic metrics designed to evaluate historical analogies across four dimensions: topic, background, process, and result. These dimensions represent the essential components of a historical event~\cite{HistoricalAnalogies}.
To measure these dimensions, we borrow the idea of \citet{jiayang-etal-2023-storyanalogy} to calculate abstract and literal similarities.
By integrating these two types of similarities across the four dimensions, our automatic evaluation metrics demonstrate a high correlation with human evaluation. 

The main contributions of this paper are summarized as follows:
\begin{itemize}[leftmargin=*]
\setlength{\itemsep}{0.1pt}
    \item To the best of our knowledge, our work is the first to explore the historical analogy in the AI community.
    \item We develop a novel, automatic multi-dimensional metric to evaluate historical analogy from a cognitive perspective, ensuring alignment with human cognition. 
    \item Through extensive experiments, we find that current LLMs have the potential for historical analogies. And by mitigating illusions and stereotypes in LLMs, our proposed self-reflection method can further improve the performance of LLMs in acquiring historical analogies.
\end{itemize}

\section{Related Work}
\label{sec:related}
\paragraph{Analogy Making}
Rooted in classical theories of analogy such as structural mapping~\cite{gentner1983structure,holyoak1996mental}, early research in the AI community primarily focuses on generating word analogies~\cite{falkenhainer1989structure,turney2005corpus,gladkova-etal-2016-analogy,fournier-etal-2020-analogies,ushio-etal-2021-bert,yuan2023analogykb} to examine the capabilities of LMs in analogy-making.
Recent advancements in LLMs~\cite{openai2022chatgpt,openai2023gpt4,llama3modelcard} have expanded this focus from simple word analogies to the generation of analogies involving more complex entities, including systems~\cite{yuan-etal-2023-beneath}, processes~\cite{bhavya-etal-2022-analogy,sultan-shahaf-2022-life,sultan2024parallelparc}, paragraphs~\cite{webb2022emergent,wijesiriwardene-etal-2023-analogical,ding2023fluid,ye2024analobench,yuan2024boosting}, measurements~\cite{chen2024beyond}, and stories~\cite{jiayang-etal-2023-storyanalogy}.
Despite these developments, most studies have concentrated on analogies within the scientific domain or everyday scenarios, overlooking the significance of historical analogy~\cite{schuman1992historical,parsons2016historical,ghilani2017looking}. 
In contrast, our research is the first to investigate and assess how LLMs can identify historical analogy, offering valuable insights for history and decision-making~\cite{HistoricalAnalogies}.

\paragraph{Language Model as Knowledge Base}
Pre-trained on extensive datasets, LLMs can implicitly encode a significant amount of knowledge within their parameters~\cite{alkhamissi2022review,xie2024adaptive,ju2024large}, enabling them to serve as Knowledge Bases (KBs)~\cite{petroni-etal-2019-language,sung-etal-2021-language,west-etal-2022-symbolic,yuan-etal-2023-distilling,xu2024survey}.
However, relying solely on LLMs for knowledge generation can lead to hallucinations~\cite{rawte2023survey,zhang2023siren,tonmoy2024comprehensive}, where the content produced seems factual but lacks grounding.
To address this issue, some researchers have proposed the retrieval-augmented generation method~\cite{shuster2021retrieval,gao2023retrieval,kirchenbauer2024hallucination} to mitigate hallucinations by leveraging external KBs. 
In this paper, we utilize LLMs to identify historical analogy, employing Wikipedia~\cite{vrandevcic2014wikidata} as an external KB to verify the authenticity of historical events and effectively mitigate hallucinations.

\section{Historical Analogy Generation}
\label{sec:pre}

\subsection{A Cognitive View for Historical Analogy}
Historical analogy compares contemporary and past situations, offering an accessible view of history and validating policies and decisions, which is a vital tool in applied history~\cite{schuman1992historical,HistoricalAnalogies,parsons2016historical}.
For example, Margaret Thatcher likened Iraq’s invasion of Kuwait to the Munich Agreement, thereby using historical analogy to support their intervention actions in Iraq~\cite{7dd02d13-40ca-3017-bcff-8c8f4247ae25}. 
In historical analogy, both events and personalities serve to formulate an argument by analogy, elucidating the present issue.
However, research conducted by historians indicates that individuals, particularly politicians, ordinarily use history badly.
They often gravitate towards the first analogy that comes to mind, are easily swayed by superficial similarities, and rarely pursue in-depth or extensive analysis~\cite{ghilani2017looking, dobney1974lessons, khong2020analogies}.
Therefore, it is crucial to develop a framework that facilitates the automatic, straightforward, and precise acquisition of historical analogy.

\subsection{Task Formulation}
Historical analogy acquisition task aims to obtain a historical event for the given event to form an analogy.  
Given the input event $\mathcal{E}_I$ and its description $\mathcal{D}_I$, the goal is to output the event from history $\mathcal{E}_H$, which is analogous to the input event. 
Figure~\ref{fig:front} presents an example of a historical analogy.

\subsection{Data Construction}\label{sec:dataset}
To comprehensively evaluate the ability of LLMs to acquire historical analogies, we categorize historical analogies into two categories, \ie, popular analogy and general analogy. 

\paragraph{Popular Analogy}
Popular analogies are analogies that are well known to the general public and already have standardized results, often proposed by newspapers, historians, and politicians, such as Figure~\ref{fig:front}.
To obtain these analogies, we manually collect samples of popular analogies from web pages and articles related to historical analogies.\footnote{The online resources are shown in Appendix~\ref{sec:resource}}
Due to the limited number of valid analogies and the presence of misuses or controversies, we end up with 20 test samples that are widely recognized, have standard answers, and show some degree of creativity.

\paragraph{General Analogy}
Since LLMs may have learned popular analogies during pre-training, we construct general analogy sets with events lacking universally recognized analogies.
Specifically, we collect 658 historical events from Google Arts and Culture.\footnote{\url{https://artsandculture.google.com/category/event}} 
These events are categorized into four themes: War, Politics, Culture and Society, and Economy.
We select 50 samples each from the first three categories and 10 from the Economy category, creating a balanced general analogical set to assess the LLM's ability to draw historical analogies across different themes.
Since there are no standardized answers for general analogies, it is necessary to develop automated evaluation metrics to assess the quality of analogies between analogy events and input events.

\subsection{Human Evaluation Metrics}
Due to the lack of quantitative criteria for evaluating historical analogies, this paper uses a ranking approach for manual assessment.
For $\mathcal{E}_I$, we employ three annotators from the history department to rank the $\mathcal{E}_H$ output from different methods according to the quality of the analogies, using a scale from 1 to n, with higher scores indicating better analogy quality.
The frequency of each method being ranked best is also calculated to assess the quality of the analogies.
Further details on the human evaluation process are provided in Appendix~\ref{appendix:Crowd-sourcing}.

\subsection{Automatic Evaluation Metrics}\label{sec:eval}
For Popular Analogies, we can calculate the \textbf{Pass@1} based on the standard answers. 
However, it is not applicable to General Analogies, necessitating the development of broader metrics for automatically evaluating historical analogies quantitatively.
Drawing on the historical applied science\footnote{\label{footnote4}\url{https://phi.history.ucla.edu/nchs/historical-thinking-standards/1-chronological-thinking/}}, we develop a multi-dimensional similarity metric (MDS) to evaluate historical analogies automatically.

\paragraph{Dimension Summary}
In historiography, the universal structure of events encompasses topic, background, process, and result.
Therefore, for an event $\mathcal{E}$ and its description $\mathcal{D}$, we utilize GPT-4 to summarize these four dimensions based on $\mathcal{D}$, resulting in $\mathcal{D}=(\mathcal{D}^{\text{Topic}}, \mathcal{D}^{\text{Background}}, \mathcal{D}^{\text{Process}}, \mathcal{D}^{\text{Result}})$.
The prompt template is shown in Appendix~\ref{appendix:prompt_eval}.

\paragraph{Multi-level Similarity}
Previous research~\cite{bunge1981analogy,jiayang-etal-2023-storyanalogy} indicates that analogies are effective when they share abstract-level similarities, such as themes, central ideas, and processes, rather than identical entities and behaviors (\ie, literal similarity).
For abstract similarity, based on the four summarized dimensions, we instruct GPT-4 to rate the abstract similarity between $\mathcal{E}_I$ and $\mathcal{E}_H$ for each dimension on a scale from 1 to 4.
The prompt template is shown in Appendix~\ref{appendix:prompt_eval}.
For literal similarity, we perform the NLTK tokenization~\cite{bird2006nltk} on each summary and calculate the Jaccard similarity~\cite{niwattanakul2013using} after removing stopwords.
A higher abstract similarity score indicates a better analogy between $\mathcal{E}_I$ and $\mathcal{E}_H$, while lower literal similarity scores indicate more innovation.
Thus, the overall multi-dimensional similarity formula is:
\vspace{-2pt}
\begin{equation}\label{eq:sx}
\begin{aligned}
MDS = \sum_{d \in \mathcal{D}} w^{d} \cdot sim_{\text{Abs}}(\mathcal{D}_I^d, \mathcal{D}_H^d) \cdot\\
max(\alpha - sim_{\text{Lit}}(\mathcal{D}_I^d, \mathcal{D}_H^d), 0), 
\end{aligned}
\end{equation}
where $\mathcal{D}=\{\text{Topic}, \text{Background}, \text{Process}, \text{Result}\}$, $w_d$ represents the weight of each dimension, $\mathcal{D}_I^d$ ($\mathcal{D}_H^d$) represents the description of $\mathcal{E}_I$ ($\mathcal{E}_H$) in the $d$ dimension.
Given that descriptions are summarized by GPT-4, even identical events may have differing descriptions. Therefore, $\alpha$ serves as a threshold to prevent overly similar analogies.

\begin{figure*}[t]
    \centering
    \includegraphics[width=0.8\linewidth]{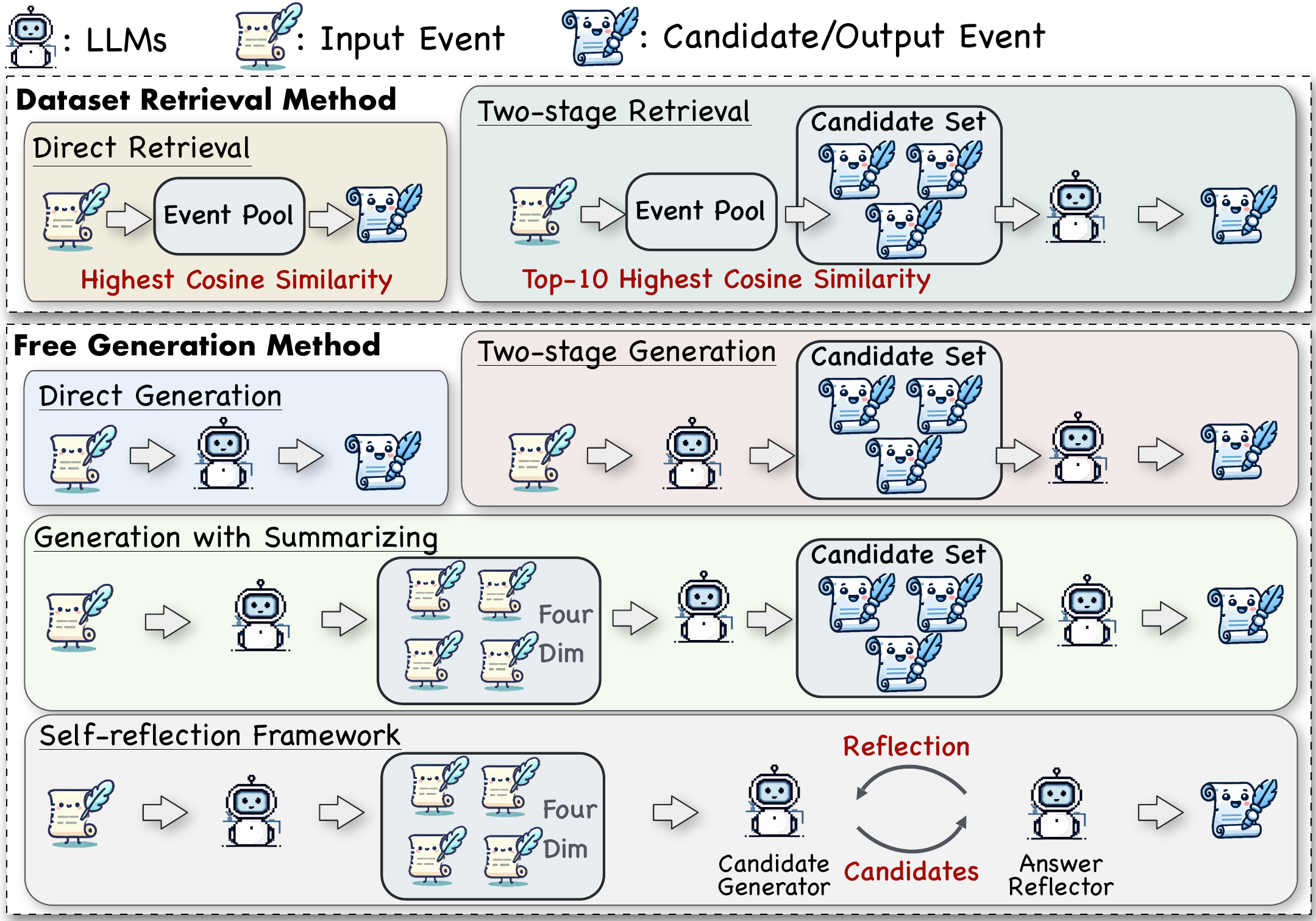}
    \caption{
    The illustration of different methods for historical analogy identification. We divide these six methods into two categories: dataset retrieval methods and free generation methods.
    }
    \label{fig:method}
\end{figure*}

\paragraph{Effectiveness of Automatic Evaluation}
To determine $w^d$ and $\alpha$, and to validate the automatic evaluation, we calculate the correlation coefficient between automatic and human evaluations.
We use GPT4 to generate four different analogies for each popular analogy as the evaluation dataset.
For the manual assessment, we employ three annotators to rank the four results, with Fleiss's $\kappa= 0.97$. 
For the automated assessment, we adopt our automatic multi-dimensional similarity metric to rank. 
For each analogy, we calculate the correlation coefficient between the two sets of rankings, then take the average across different analogies to obtain the final correlation coefficient.

The results show that the best correlation coefficient with the manual results is obtained when the dimension weights are (0.5, 1, 2, 2) and the similarity threshold $\alpha$ is 0.35. 
In this setting, the Kappa coefficient~\cite{cohen1960coefficient} is 0.67, and the Pearson~\cite{pearson1920notes} and Spearman correlation coefficients~\cite{spearman1961proof} are 0.72 and 0.73, confirming the reliability of automatic evaluation.

\section{Method}
\label{sec:method}
In this section, we explore various methods of leveraging LLMs to get historical analogies.  
These methods fall into two categories:
1) \textit{dataset retrieval methods} 
and 2) \textit{free generation methods}. 
The illustration of these methods is shown in Figure~\ref{fig:method}.
The prompt templates for LLMs in each method are shown in the Appendix~\ref{appendix:prompt_method}.

\subsection{Dataset Retrieval Method}
A common practice to obtain analogous events is to select from existing datasets. In this paper, we use Google Arts and Culture as an event pool for LLMs to retrieve suitable analogies for a specified event. 
We implement two retrieval strategies:

\paragraph{Direct Retrieval}
This method embeds the description of the given event and events in the pool using \texttt{text-embedding-3-small}~\cite{neelakantan2022text}. The event with the highest cosine similarity is then selected as $\mathcal{E}_H$.

\paragraph{Two-stage Retrieval}
This method first selects the top 10 historical events from the event pool using cosine similarity.
Then, given their descriptions, LLM is asked to select the most appropriate analogies from the candidate set. 

\newcolumntype{a}{>{\columncolor{BlueGreen!10}\centering\arraybackslash}p{0.88cm}}
\newcolumntype{b}{>{\columncolor{Blue!10}\centering\arraybackslash}p{0.88cm}}
\newcolumntype{d}{>{\columncolor{Purple!10}\centering\arraybackslash}p{0.88cm}}
\newcolumntype{q}{>{\columncolor{Gray!10}\centering\arraybackslash}p{0.88cm}}
\newcolumntype{e}{>{\columncolor{Brown!10}\centering\arraybackslash}p{0.88cm}}

\setlength\tabcolsep{1.3pt}
\begin{table*}[t]
  \centering
  \caption{Results of different methods on Popular Analogies and General Analogies based on ChatGPT and Llama3.1-8B. ``Abs'' (``Lit'') denotes abstract similarity (literal similarity). ``T'', ``B'', ``P'', ``R'' denote the dimensions of Topic, Background, Process and Result. ``MDS'' denotes multi-dimensions similarity. The best results are \textbf{bolded}, and the second best ones are \uline{underlined}, both counted to four decimal places.
  }
    \small
    \begin{tabular}{claaabbbdddqqqe}
    \toprule
    \textbf{Dataset} & \multicolumn{1}{c}{\textbf{Method}} & \textbf{T$_{\texttt{Abs}}$} & \textbf{T$_{\texttt{Lit}}$} & \textbf{T$_{\texttt{All}}$}  & \textbf{B$_{\texttt{Abs}}$} & \textbf{B$_{\texttt{Lit}}$} & \textbf{B$_{\texttt{All}}$}  & \textbf{P$_{\texttt{Abs}}$} & \textbf{P$_{\texttt{Lit}}$} & \textbf{P$_{\texttt{All}}$}  & \textbf{R$_{\texttt{Abs}}$} & \textbf{R$_{\texttt{Lit}}$} & \textbf{R$_{\texttt{All}}$}  & \textbf{MDS} \\
    \midrule
    \multirow{14}[8]{*}{Popular} & \multicolumn{13}{c}{GPT-3.5-Turbo} \\
\cmidrule{2-15}
& Direct Re.  & 2.70  & 0.15  & 0.54  & 2.55  & 0.14  & 0.61  & 2.70  & \underline{0.09}  & 0.70  & 2.70  & \underline{0.09}  & 0.70  & 3.67  \\
          & Two-stage Re.  & 2.85  & \textbf{0.13} & 0.59  & 2.45  & 0.12  & 0.53  & 2.65  & \textbf{0.08} & 0.69  & 2.60  & 0.10  & 0.60  & 3.41  \\
          \cdashlinelr{2-15}
          & Direct Gen.   & 3.10  & 0.15  & 0.64  & 2.64  & 0.11  & \underline{0.68}  & 2.94  & 0.12  & 0.73  & \underline{3.15}  & 0.10  & 0.75  & 3.97  \\
          & Two-stage Gen.   & 3.25  & 0.16  & 0.65  & \textbf{2.90} & 0.12  & 0.67  & 2.80  & 0.13  & 0.68  & 2.80  & 0.12  & 0.69  & 3.74  \\
          & Summarizing  & \underline{3.30}  & 0.14  & \underline{0.67}  & 2.70  & \underline{0.11}  & 0.63  & \textbf{3.30} & 0.10  & \textbf{0.82} & \textbf{3.19} & 0.10  & \underline{0.76}  & \underline{4.14}  \\
          & Self-reflection   & \textbf{3.40} & \underline{0.13}  & \textbf{0.71} & \underline{2.89}  & \textbf{0.09} & \textbf{0.73} & \underline{3.09}  & 0.10  & \underline{0.75}  & 3.09  & \textbf{0.09} & \textbf{0.79} & \textbf{4.18} \\
\cmidrule{2-15}          
& \multicolumn{13}{c}{Llama3.1-8B} \\
\cmidrule{2-15}         
& Direct Re.     & 2.70  & 0.15  & 0.54  & 2.55  & 0.14  & 0.61  & 2.70  & 0.09  & 0.70  & 2.70  & \textbf{0.09} & 0.70  & 3.67  \\
          & Two-stage Re.& 2.80  & \textbf{0.11} & 0.63  & 2.45  & 0.10  & 0.59  & 2.60  & \underline{0.08}  & 0.69  & 2.44  & 0.12  & 0.55  & 3.38  \\
          \cdashlinelr{2-15}
          & Direct Gen.   & \underline{3.30}  & 0.13  & \textbf{0.70} & 2.69  & 0.09  & 0.69  & \underline{2.90}  & 0.10  & 0.69  & \textbf{3.10} & 0.10  & \textbf{0.74} & 3.90  \\
          & Two-stage Gen.  & 2.94  & \underline{0.13}  & 0.64  & 2.55  & \underline{0.08}  & 0.67  & 2.80  & \textbf{0.08} & \underline{0.73}  & 2.80  & 0.10  & 0.68  & 3.81  \\
          & Summarizing & 3.24  & 0.14  & 0.64  & \underline{2.74}  & \textbf{0.08} & \underline{0.74}  & 2.69  & 0.08  & 0.70  & \underline{2.94}  & \underline{0.09}  & \underline{0.73}  & \underline{3.92}  \\
          & Self-reflection   & \textbf{3.34} & 0.13  & \underline{0.70}  & \textbf{2.84} & 0.08  & \textbf{0.74 } & \textbf{3.15} & 0.09  & \textbf{0.81} & 2.89  & 0.10  & 0.71  & \textbf{4.13} \\
    \midrule
    \multirow{14}[8]{*}{General} & \multicolumn{13}{c}{GPT-3.5-Turbo} \\
\cmidrule{2-15}         
& Direct Re.  & 3.29  & 0.18  & 0.53  & 3.00  & 0.13  & 0.67  & 2.97  & \underline{0.11}  & 0.69  & 2.99  & 0.12  & 0.66  & 3.64  \\
          & Two-stage Re.  & 2.93  & 0.19  & 0.51  & 2.69  & 0.15  & 0.58  & 2.63  & 0.12  & 0.60  & 2.75  & 0.14  & 0.58  & 3.21  \\
         \cdashlinelr{2-15}
          & Direct Gen.  & 2.88  & \textbf{0.13} & \underline{0.62}  & 2.67  & \textbf{0.10} & 0.65  & 2.63  & \textbf{0.09} & 0.69  & 2.79  & \textbf{0.10} & \underline{0.68}  & 3.69  \\
          & Two-stage Gen.  & 3.20  & 0.20  & 0.57  & 2.82  & 0.16  & 0.63  & 3.01  & 0.13  & 0.70  & 2.99  & 0.13  & 0.67  & 3.65  \\
          & Summarizing  & \underline{3.49}  & 0.18  & \textbf{0.64} & \underline{3.02}  & 0.13  & \underline{0.68}  & \underline{3.11}  & 0.12  & \underline{0.74}  & \underline{3.07}  & 0.13  & 0.67  & \underline{3.83}  \\
          & Self-reflection  & \textbf{3.52} & \underline{0.17}  & 0.61  & \textbf{3.21} & \underline{0.12}  & \textbf{0.73} & \textbf{3.16} & 0.11  & \textbf{0.75} & \textbf{3.13} & \underline{0.12}  & \textbf{0.70} & \textbf{3.93} \\
\cmidrule{2-15}          & \multicolumn{13}{c}{Llama3.1-8B} \\
\cmidrule{2-15}         
& Direct Re.  & 3.29  & 0.18  & 0.53  & 3.00  & 0.13  & 0.67  & 2.97  & 0.11  & 0.69  & 2.99  & 0.12  & 0.66  & 3.64  \\
          & Two-stage Re.  & 3.11  & \underline{0.16}  & 0.58  & 2.78  & \textbf{0.11} & 0.64  & 2.81  & \textbf{0.09} & 0.71  & 2.73  & \underline{0.11}  & 0.63  & 3.60  \\
          \cdashlinelr{2-15}
          & Direct Gen.  & \underline{3.44}  & 0.19  & 0.60  & \underline{3.08}  & 0.15  & 0.67  & \underline{3.04}  & 0.13  & 0.72  & 3.01  & 0.14  & 0.66  & 3.73  \\
          & Two-stage Gen.  & 3.21  & \textbf{0.15} & \textbf{0.63} & 2.91  & \underline{0.11}  & 0.69  & 2.91  & \underline{0.10}  & 0.73  & 2.80  & \textbf{0.11} & 0.66  & 3.77  \\
          & Summarizing  & 3.41  & 0.17  & 0.61  & \textbf{3.11} & 0.11  & \textbf{0.72} & 3.02  & 0.10  & \underline{0.74}  & \underline{3.09}  & 0.12  & \underline{0.70}  & \underline{3.90}  \\
          & Self-reflection   & \textbf{3.46} & 0.18  & \underline{0.62}  & \underline{3.08}  & 0.13  & \underline{0.70}  & \textbf{3.08} & 0.11  & \textbf{0.75} & \textbf{3.12} & 0.13  & \textbf{0.70 } & \textbf{3.91 } \\
    \bottomrule
    \end{tabular}%
  \label{tab:main-result}%
\end{table*}%
\subsection{Free Generation Method}
\label{lab:freegeneration}

Due to the growing number of historical events, relying on a fixed dataset for analogies can lead to issues such as high overhead, slow processing, and challenges in updating. 
Since LLMs have learned extensive knowledge about historical events during pre-training, we can employ LLMs to generate analogous historical events.

\paragraph{Direct Generation}
Given $\mathcal{E}_I$ and $\mathcal{D}_I$, this method directly asks LLMs to generate the analogous historical event.
However, it heavily depends on LLMs' knowledge and can be easily influenced by biases and stereotypes from pre-training.

\paragraph{Two-stage Generation}
Direct generation can lead to suboptimal results and even produce fictional historical events with hallucinations.
To achieve a broader exploration, we ask LLMs to propose 10 candidate events based on $\mathcal{D}_I$. 
Given the potential for hallucination, each candidate must be verified through Wikipedia to confirm its authenticity.
Then, LLMs compare the $\mathcal{D}_I$ and the descriptions of candidate events retrieved from Wikipedia, selecting the most appropriate one as $\mathcal{E}_H$.

\paragraph{Generation with Summarizing}
As mentioned in $\mathsection$~\ref{sec:eval}, the common structure of events encompasses topic, background, process, and result.
Thus, we can ask LLMs to summarize the input and candidates into these four dimensions and combine the summaries to form new descriptions to participate in the steps of the two-stage generation.
Compared with the original descriptions, the summaries have shorter lengths and more effective information, so LLM can better understand the events and compare the similarities and differences in different dimensions to obtain better analogical results.

\paragraph{Self-reflection Framework}
Based on the evaluation results in $\mathsection$~\ref{sec:result}, the generation with summarizing improves output quality but is prone to stereotyping when proposing candidates, and the limitation of only 10 candidates restricts LLM options.
Research on the self-reflection method~\cite{shinn2023reflexion, renze2024self, wang2024theoretical} shows that LLMs can provide feedback and update the unsuitable candidates.
Inspired by this, we design two LLM-based modules: the Candidate Generator and the Answer Reflector.
These modules collaboratively generate historical analogies through iterative processes.
In each iteration, Candidate Generator proposes five candidates based on the $\mathcal{E}_I$'s descriptions of four dimensions. 
The Answer Reflector then assesses the candidate set. 
If no candidates are suitable as analogous historical events, the Answer Reflector instructs the Candidate Generator to revise the candidate set for the next iteration. 
If a suitable candidate is found, the Answer Reflector outputs $\mathcal{E}_H$ and concludes the iteration.
Additionally, we also verify each candidate through Wikipedia to confirm its authenticity.

\section{Results}
\label{sec:result}
This section evaluates methods for historical analogy acquisition and identifies core challenges, such as stereotypes and differing perspectives.
Furthermore, ablation studies reveal the critical components of the framework and validate the potential of LLMs for this task.

\subsection{Model Choice}
We use the open-source model Llama3.1-8B-Instruct~\cite{llama3modelcard} and the closed-source model gpt-3.5-turbo-0125~\cite{openai2022chatgpt} for the main experiment, with the temperature set to 0.1.

\subsection{Main Result}
\begin{figure}[t]
    \centering
    \includegraphics[width=\linewidth]{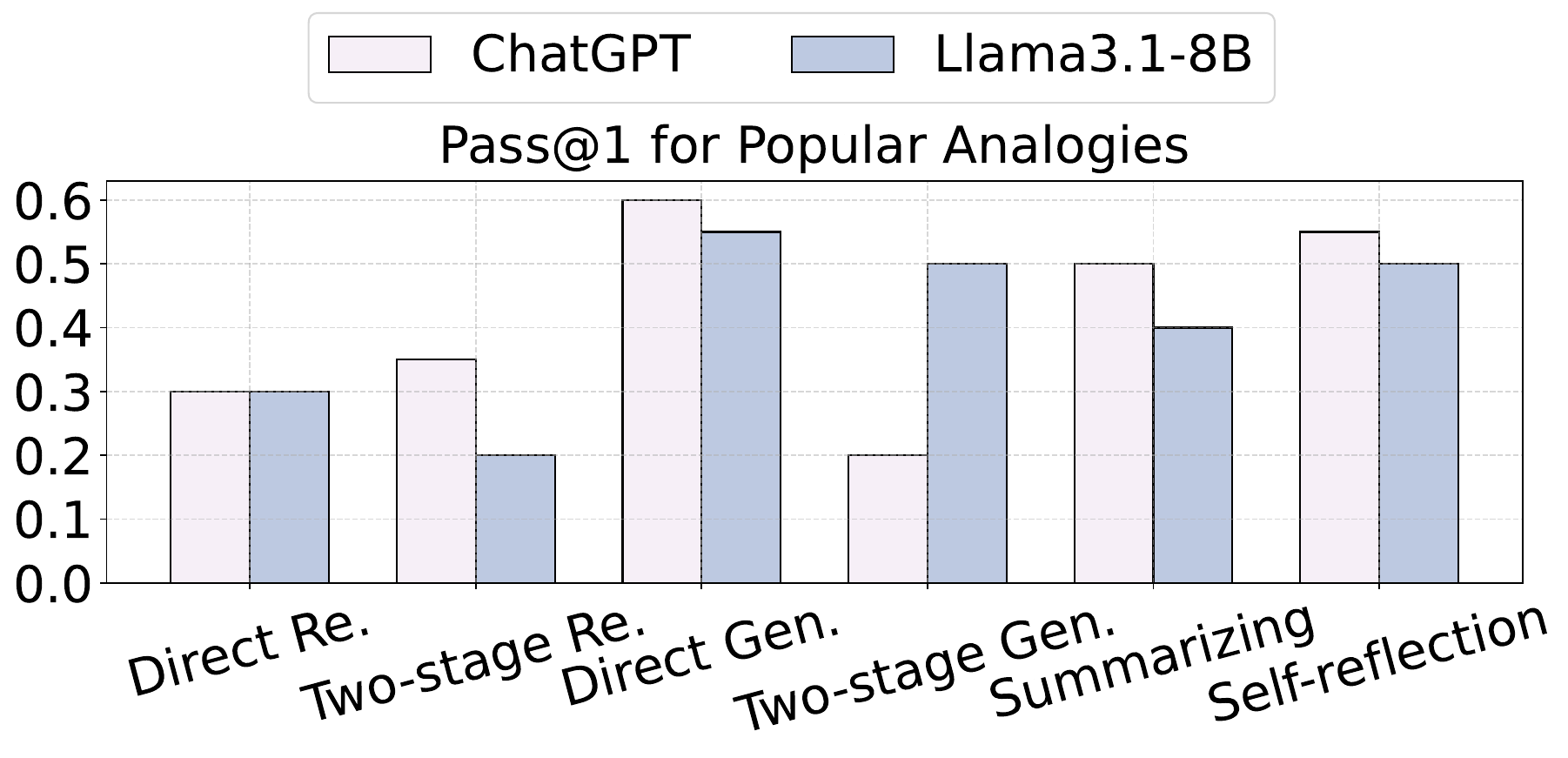}
    \caption{Pass@1 results of different methods on the Popular Analogies.}
    \label{fig:passk}
\end{figure}

\paragraph{Automatic Evaluation Results}
The results are shown in Table~\ref{tab:main-result} and Figure~\ref{fig:passk}.
And Appendix~\ref{appendix:stability} provides confidence intervals for the results.
We find that:
\begin{inparaenum}[\it 1)]
\item Both Llama and ChatGPT perform better on the Popular Analogy than on the General Analogy. 
In particular, Direct Generation method achieves a high Pass@1 in Popular Analogy.
This discrepancy suggests potential data leakage during the pre-training phase within the Popular Analogy, emphasizing the importance of including General Analogy in evaluations.
\item Free generation methods outperform dataset retrieval methods significantly, with an average improvement of 0.25. 
This improvement likely arises because a finite dataset cannot encompass the vast expanse of historical data, making generation from LLMs preferable to retrieval for historical analogies.
\item The self-reflection method achieves the highest results for both open and closed-source models, indicating that incorporating reflection with feedback can enhance the quality of analogies.
\item The summarizing method demonstrates notable enhancements over two-stage generation across both models and datasets, highlighting the effectiveness of dimension splitting in improving historical analogy generation.
\item Surprisingly, the two-stage method generally underperforms compared to the direct method. This may be attributed to the lengthy and detailed descriptions, making it more difficult for the LLM to choose during the selection process.
\end{inparaenum}

\begin{figure}[t]
    \centering
    \includegraphics[width=\linewidth]{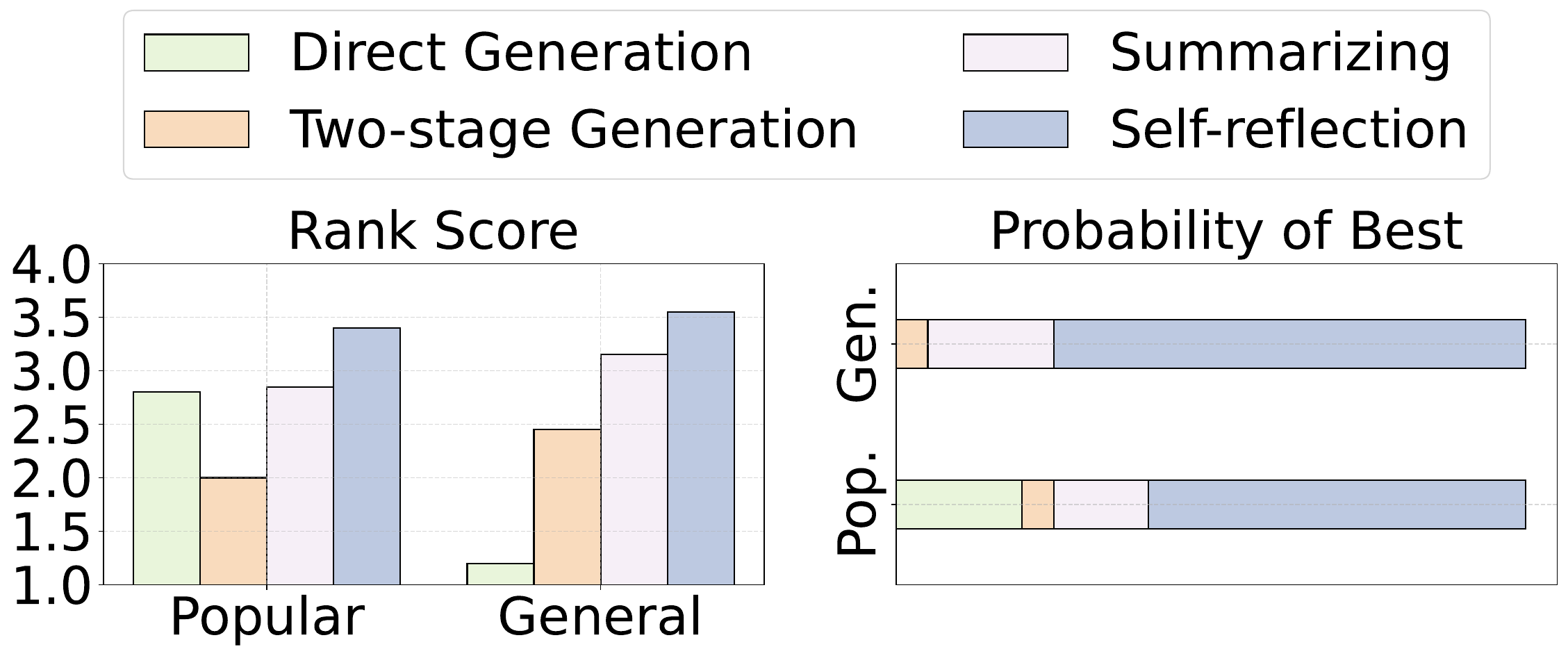}
    \caption{Human evaluation results for free generation methods.
    }
    \label{fig:human-eval}
\end{figure}

\begin{figure}[t]
    \centering
    \includegraphics[width=\linewidth]{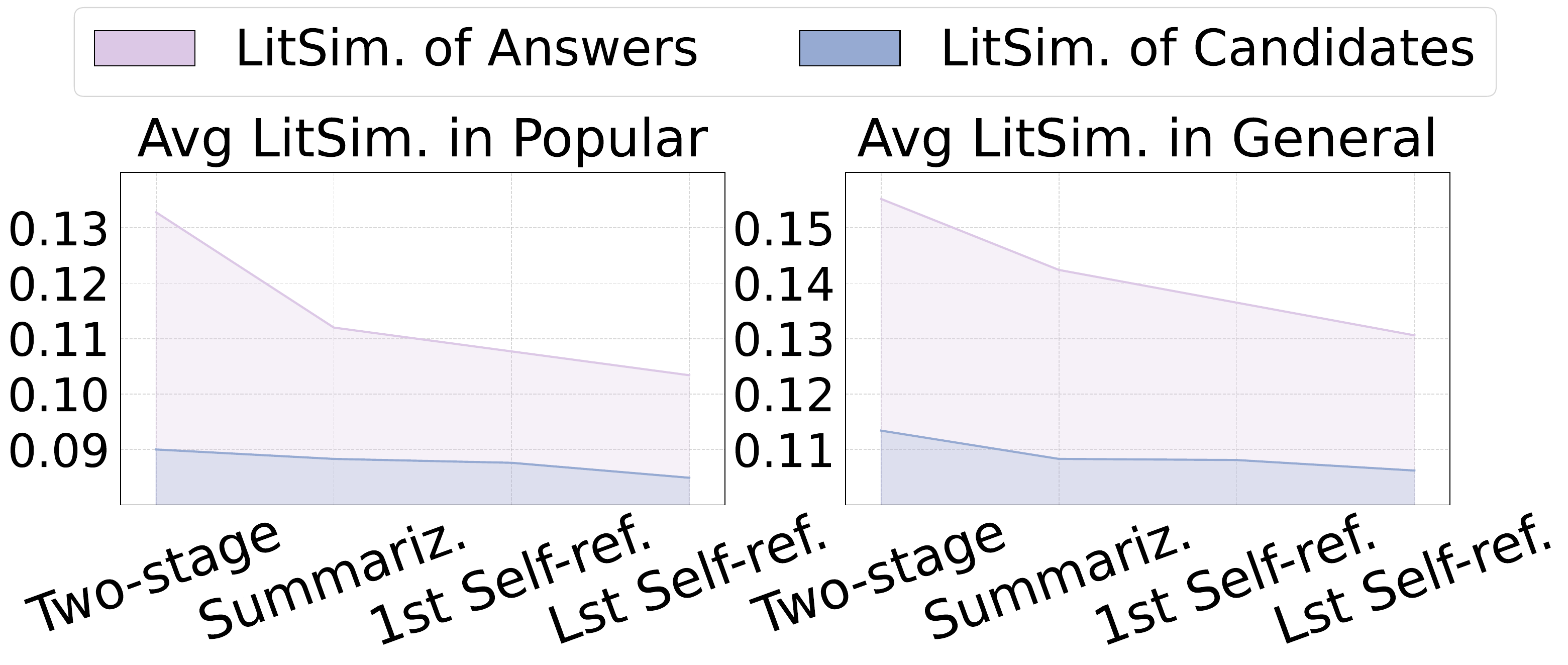}
    \caption{Average literal similarity of candidates and answers for free generation method. ``1st'' indicates the first round of self-reflection and ``Lst' is the last round.
    }
    \label{fig:stereotypes}
\end{figure}

\newcolumntype{a}{>{\columncolor{BlueGreen!10}\centering\arraybackslash}p{0.9cm}}
\newcolumntype{b}{>{\columncolor{Blue!10}\centering\arraybackslash}p{0.9cm}}
\newcolumntype{d}{>{\columncolor{Purple!10}\centering\arraybackslash}p{0.9cm}}
\newcolumntype{q}{>{\columncolor{Gray!10}\centering\arraybackslash}p{0.9cm}}
\newcolumntype{e}{>{\columncolor{Brown!10}\centering\arraybackslash}p{0.9cm}}
\newcolumntype{f}{>{\centering\arraybackslash}p{0.9cm}}
\setlength\tabcolsep{2pt}
\begin{table}[t]
  \centering
  \caption{Results of different candidate set sizes in the self-reflection method. \textbf{Ref.} indicates the average number of times the reflection is performed.}
    \small
    \begin{tabular}{lfabdqe}
    \toprule
    \#\textbf{Can.} & \textbf{Ref.} & \textbf{T$_{\texttt{All}}$}  & \textbf{B$_{\texttt{All}}$}  & \textbf{P$_{\texttt{All}}$}  & \textbf{R$_{\texttt{All}}$}  & \textbf{MDS} \\
    \midrule
    C=1 & 0.10  & 0.60  & 0.70  & 0.73  & 0.70  & 3.86  \\
    C=3 & 0.14  & 0.61  & 0.68  & 0.74  & 0.71  & 3.87  \\
    C=5 & 0.11  & 0.61  & 0.73  & 0.75  & 0.70  & 3.93  \\
    C=10 & 0.09  & 0.63  & 0.70  & 0.74  & 0.70  & 3.89  \\
    C=15 & 0.09  & 0.61  & 0.72  & 0.77  & 0.69  & 3.95  \\
    \bottomrule
    \end{tabular}%
  \label{tab:candidate-num}%
\end{table}%

\newcolumntype{a}{>{\columncolor{BlueGreen!10}\centering\arraybackslash}p{0.9cm}}
\newcolumntype{b}{>{\columncolor{Blue!10}\centering\arraybackslash}p{0.9cm}}
\newcolumntype{d}{>{\columncolor{Purple!10}\centering\arraybackslash}p{0.9cm}}
\newcolumntype{q}{>{\columncolor{Gray!10}\centering\arraybackslash}p{0.9cm}}
\newcolumntype{e}{>{\columncolor{Brown!10}\centering\arraybackslash}p{0.9cm}}
\setlength\tabcolsep{2pt}
\begin{table}[t]
  \centering
  \caption{Results for different warmup turns in the self-reflection method.}
    \small
    \begin{tabular}{labdqe}
    \toprule
    \textbf{Warmup} & \textbf{T$_{\texttt{All}}$}  & \textbf{B$_{\texttt{All}}$}  & \textbf{P$_{\texttt{All}}$}  & \textbf{R$_{\texttt{All}}$}  & \textbf{MDS} \\
    \midrule
    W=0 & 0.61  & 0.73  & 0.75  & 0.70  & 3.93  \\
    W=2 & 0.66  & 0.72  & 0.77  & 0.73  & 4.03  \\
    W=5 & 0.63  & 0.70  & 0.73  & 0.68  & 3.85  \\
    W=10 & 0.66  & 0.70  & 0.75  & 0.71  & 3.94  \\
    \bottomrule
    \end{tabular}%
  \label{tab:warmup}%
\end{table}%

\paragraph{Human Evaluation Results}
To further assess the performance of each method, we conduct a manual evaluation of the four free generation methods based on ChatGPT. 
The results are presented in Figure~\ref{fig:human-eval}. 
In alignment with the automated results, the self-reflection method receive the highest ranking score and the highest percentage of optimal. 
Also, due to the internal knowledge leakage in LLM, direct generation performs well in Popular Analogies but poorly in General Analogies.

\subsection{Detailed Analysis}

\begin{figure*}[t]
    \centering
    \includegraphics[width=\linewidth]{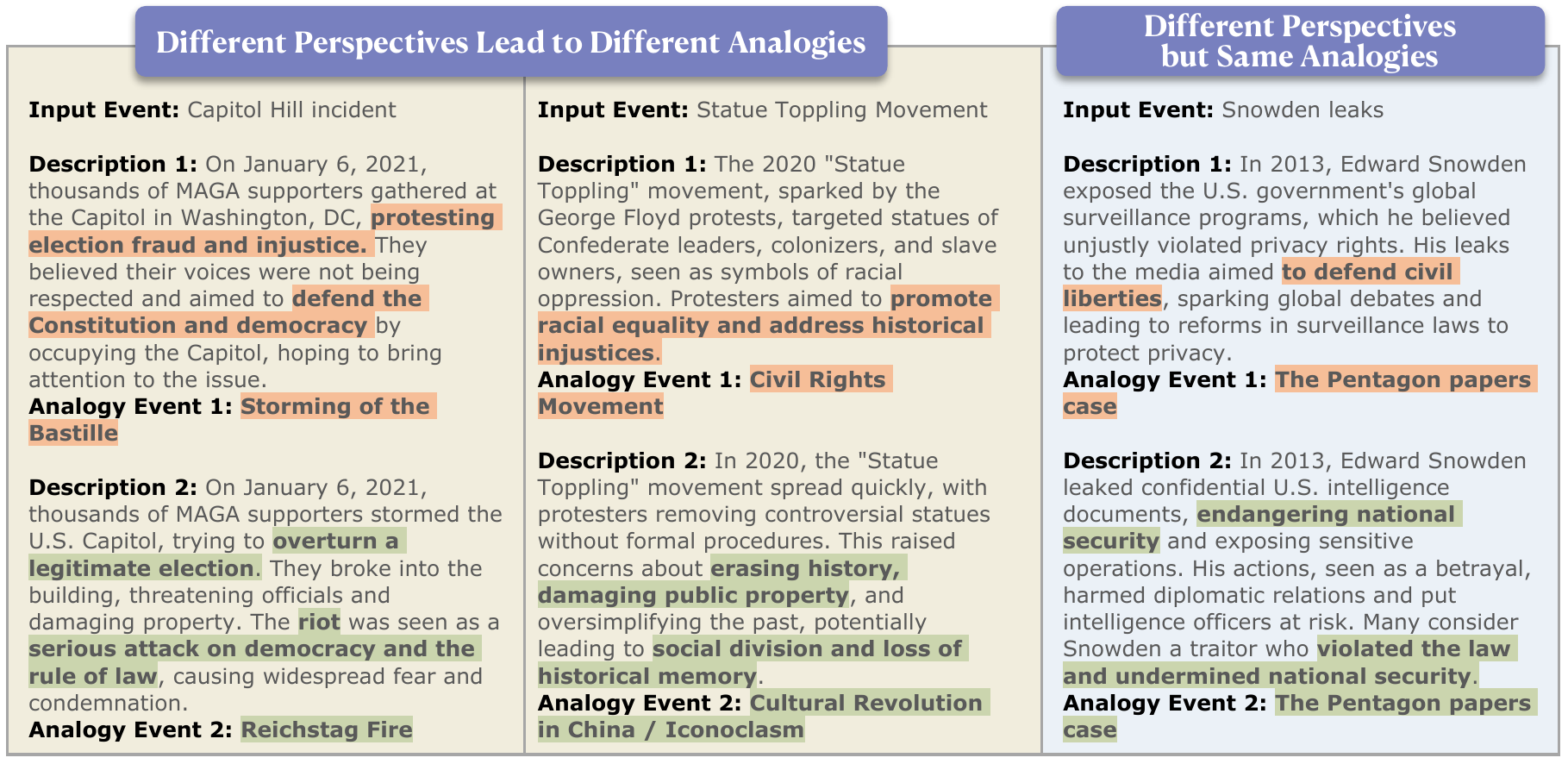}
    \caption{Case studies of historical analogy from different perspectives. Different perspectives often lead to distinct analogies, although a few analogies remain the same due to the ability to interpret the results from multiple viewpoints.
    }
    \label{fig:case_study}
\end{figure*}

\begin{figure}[t]
    \centering
    \includegraphics[width=\linewidth]{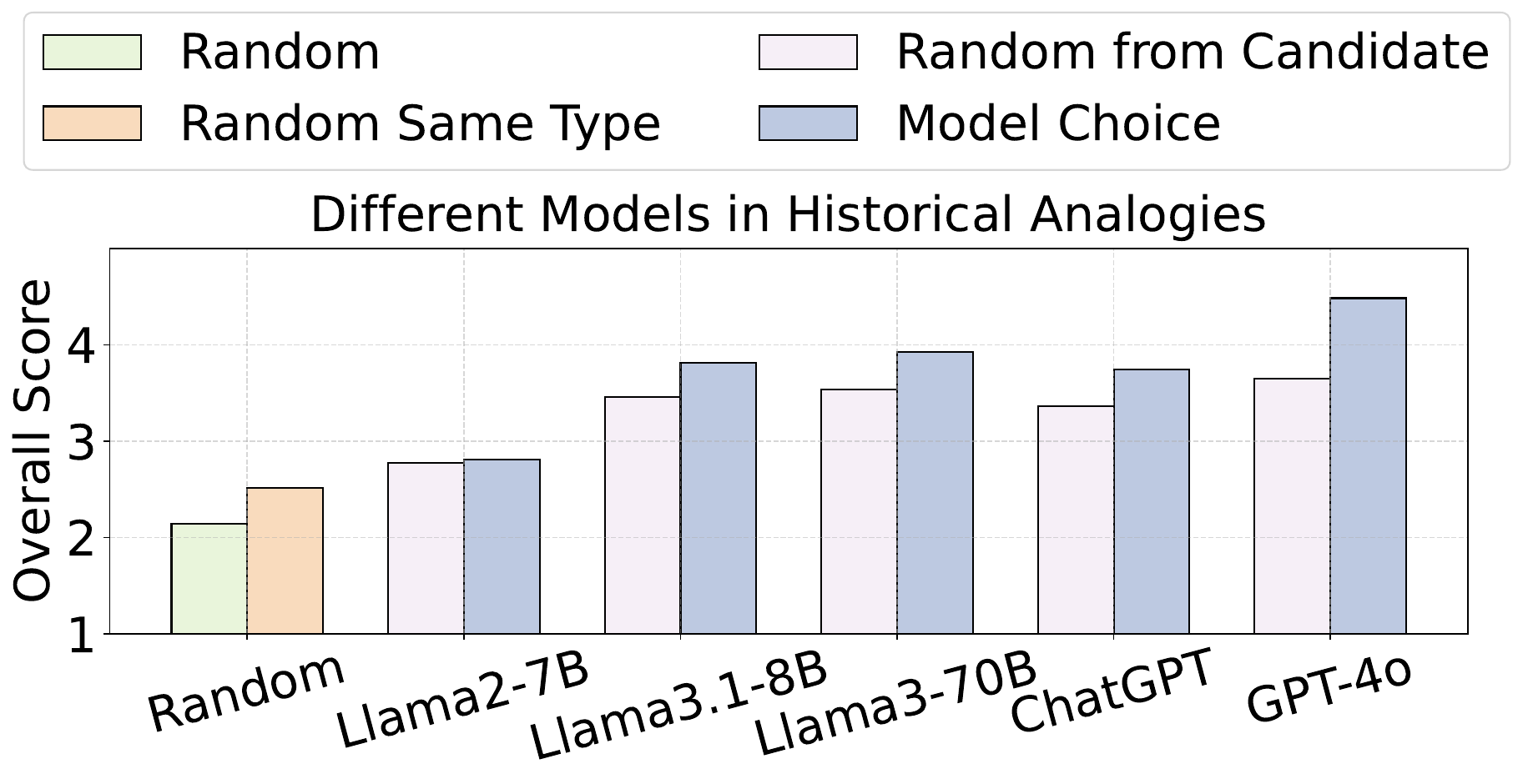}
    \caption{Performance of different models in historical analogies, including proposing candidate sets and selecting analogous event.
    }
    \label{fig:different-model}
\end{figure}

\paragraph{Stereotypes in Historical Analogies}
Stereotypes in historical analogies are mainly manifested in generating events that focus on the same entities, \eg countries, people, rather than on core ideas.
In order to analyse the impact of stereotypes, we count the literal similarity scores of candidates and answers generated by the free generation method, which reflects the degree of stereotyping by measuring the proportion of shared entities.
Figure~\ref{fig:stereotypes} shows the results, which align with the motivations in $\mathsection$~\ref{lab:freegeneration}.
In both Popular and General Analogy, the candidates and answers of the self-reflection method show the least stereotyping and further decrease with iteration.

\paragraph{Candidate Number and Reflective Rounds}
To further explore self-reflection method, we first test the effect of different candidate set sizes on it.
The results in Table~\ref{tab:candidate-num} show that increasing the candidate set size improves performance, but gains plateau after five candidates while token consumption continues to rise.

However, the low Ref. shows that only about 10\% data executed reflection, indicating that LLMs prefer to accept the current candidate, even when the candidate set is small.
To explore the impact of the number of reflection rounds on performance, we further require the LLM to warm up with a few rounds of reflection without output. 
Table~\ref{tab:warmup} reveals that while a few warmups can slightly improve performance, additional rounds do not continue this trend and may reduce effectiveness due to inappropriate reflection.

\paragraph{Proposing and Selection Capability of LLMs}
As described in $\mathsection$~\ref{sec:method}, the methods of two-stage generation, summarizing and self-reflection need to propose a candidate set and select the analogous event $\mathcal{E}_H$ from this set.
To prove the effectiveness of LLMs in proposing an appropriate candidate set and selecting the $\mathcal{E}_H$ from the set, we design the three ablated variants in the two-stage generation method for exploration:
\begin{inparaenum}[\it 1)]
\item random selection from the event pool;
\item random selection from the event pool within the same theme;
\item random selection from the candidate set proposed by LLMs.
\end{inparaenum}

The results are shown in Figure~\ref{fig:different-model}.
The evidence provided by \textcolor{color1}{\rule{1.5ex}{1.5ex}} < \textcolor{color3}{\rule{1.5ex}{1.5ex}} and \textcolor{color2}{\rule{1.5ex}{1.5ex}} < \textcolor{color3}{\rule{1.5ex}{1.5ex}} confirms that all models, ranging from smaller ones like Llama2-7B~\cite{touvron2023llama} to larger ones like GPT-4~\cite{openai2023gpt4}, are capable of generating candidate sets for historical analogies. 
Furthermore, stronger LLMs, \eg, ChatGPT, GPT-4, demonstrate superior selection performance, as indicated by \textcolor{color3}{\rule{1.5ex}{1.5ex}} < \textcolor{color4}{\rule{1.5ex}{1.5ex}}, showing their effectiveness in selecting the historical event analogous to the input event.
However, Llama2-7B shows limited improvement over random selection in generating historical analogies, suggesting that there is room for enhancing the general capabilities of LLMs in this domain.

\paragraph{Historical Analogies from Different Perspectives}
Different individuals may describe the same event in various ways. 
We are interested in determining whether these differing perspectives influence the historical analogies generated from LLMs. 
To investigate this, we select several controversial events, manually create descriptions from different viewpoints, and utilize a self-reflection method based on ChatGPT to generate historical analogies.
Figure~\ref{fig:case_study} presents some typical cases. 
Our findings indicate that varying descriptions can indeed lead to different analogical outcomes. 
For instance, the Capitol Hill incident might be analogous to the Storming of the Bastille from the Republican Party's perspective or to the Reichstag fire from the Democrats' perspective. 
However, different descriptions may also produce the same analogies, since analogous events can also have diverse interpretations.
Future research could focus on developing methods to identify and evaluate historical analogies based on diverse perspectives.

\section{Conclusion}
\label{sec:conclusion}
In this paper, we explore the concept of historical analogy and examine the ability of LLMs to acquire historical analogies for given events. 
We create an automatic multi-dimensional similarity metrics to fairly assess the quality of historical analogies, and perform numerous experiments with different models, which show that LLMs have the potential for historical analogies.
In addition, we design an optimization method, self-reflection, which breaks from the stereotypes through multiple rounds of reflection and improves the historical analogical performance of the model.

\section*{Limitations}
\label{sec:limitation}
First, our evaluation mainly focuses on the accuracy of the analogous historical events, without assessing the reasons provided by the model due to the challenges in automatic evaluation of reasoning.
Second, while our evaluation considers four specific dimensions to determine the correctness of historical analogies, it is important to note that in real-life contexts, additional factors such as gender, party affiliation, and motivation might also be considered, particularly by politicians.
However, we believe that the evaluation of these additional dimensions could be automated through the application of our proposed evaluation methodology.
Although we include historical analogies from various perspectives, assessing the rationality and applicability of these analogies across different perspectives remains challenging.

\section*{Ethics Statement}
\label{sec:Ethics}
We hereby acknowledge that all authors of this work are aware of the provided ACL Code of Ethics and honor the code of conduct.

\paragraph{Use of Human Annotations} Evaluation on the identified historical analogies from LLMs is implemented by three annotators recruited by our institution. The construction team remains anonymous to the authors. 
We ensure that the privacy rights of all annotators are respected throughout the annotation process. 
All annotators are compensated above the local minimum wage and consent to the use of these historical analogies for research purposes, as described in our paper. 
The annotation details are shown in Appendix~\ref{appendix:Crowd-sourcing}.

\paragraph{Risks}
The analogy sets used in the experiment, including the popular and general sets, are derived from publicly accessible sources.
We have reviewed these analogies to ensure they are free from socially harmful or toxic language. 
However, we cannot guarantee that they will not offend certain groups. 
Furthermore, evaluating historical analogies depends on common sense, and individuals from diverse backgrounds may have different perspectives. 
We use ChatGPT~\cite{openai2022chatgpt} to correct grammatical errors in this paper.

\section*{Acknowledgements}

This work was supported by funding from Ant Group.

\bibliography{custom}

\begin{thebibliography}{64}
\providecommand{\natexlab}[1]{#1}

\bibitem[{Achenbaum(1983)}]{achenbaum1983making}
W~Andrew Achenbaum. 1983.
\newblock The making of an applied historian: Stage two.
\newblock \emph{The Public Historian}, 5(2):21--46.

\bibitem[{AI-Meta(2024)}]{llama3modelcard}
AI-Meta. 2024.
\newblock \href {https://github.com/meta-llama/llama3/blob/main/MODEL_CARD.md} {Llama 3 model card}.

\bibitem[{AlKhamissi et~al.(2022)AlKhamissi, Li, Celikyilmaz, Diab, and Ghazvininejad}]{alkhamissi2022review}
Badr AlKhamissi, Millicent Li, Asli Celikyilmaz, Mona Diab, and Marjan Ghazvininejad. 2022.
\newblock A review on language models as knowledge bases.
\newblock \emph{arXiv preprint arXiv:2204.06031}.

\bibitem[{Axelrod and Forster(2017)}]{AXELROD20178}
Robert Axelrod and Larissa Forster. 2017.
\newblock \href {https://doi.org/10.1016/j.rie.2016.08.001} {How historical analogies in newspapers of five countries make sense of major events: 9/11, mumbai and tahrir square}.
\newblock \emph{Research in Economics}, 71(1):8--19.

\bibitem[{Bartha(2013)}]{bartha2013analogy}
Paul Bartha. 2013.
\newblock Analogy and analogical reasoning.

\bibitem[{Bhavya et~al.(2024)Bhavya, Sehgal, Xiong, and Zhai}]{bhavya-etal-2024-anade1}
Bhavya Bhavya, Shradha Sehgal, Jinjun Xiong, and ChengXiang Zhai. 2024.
\newblock \href {https://aclanthology.org/2024.eacl-long.103} {{A}na{DE}1.0: A novel data set for benchmarking analogy detection and extraction}.
\newblock In \emph{Proceedings of the 18th Conference of the European Chapter of the Association for Computational Linguistics (Volume 1: Long Papers)}, pages 1723--1737, St. Julian{'}s, Malta. Association for Computational Linguistics.

\bibitem[{Bhavya et~al.(2022)Bhavya, Xiong, and Zhai}]{bhavya-etal-2022-analogy}
Bhavya Bhavya, Jinjun Xiong, and ChengXiang Zhai. 2022.
\newblock \href {https://aclanthology.org/2022.inlg-main.25} {Analogy generation by prompting large language models: A case study of instructgpt}.
\newblock In \emph{Proceedings of the 15th International Conference on Natural Language Generation}, pages 298--312, Waterville, Maine, USA and virtual meeting. Association for Computational Linguistics.

\bibitem[{Bird(2006)}]{bird2006nltk}
Steven Bird. 2006.
\newblock Nltk: the natural language toolkit.
\newblock In \emph{Proceedings of the COLING/ACL 2006 Interactive Presentation Sessions}, pages 69--72.

\bibitem[{Bunge(1981)}]{bunge1981analogy}
Mario Bunge. 1981.
\newblock Analogy between systems.
\newblock \emph{International Journal Of General System}, 7(4):221--223.

\bibitem[{Chen et~al.(2022)Chen, Xu, Fu, Shi, Li, Zhang, Sun, Li, Xiao, and Zhou}]{chen-etal-2022-e}
Jiangjie Chen, Rui Xu, Ziquan Fu, Wei Shi, Zhongqiao Li, Xinbo Zhang, Changzhi Sun, Lei Li, Yanghua Xiao, and Hao Zhou. 2022.
\newblock \href {https://doi.org/10.18653/v1/2022.findings-acl.311} {{E}-{KAR}: A benchmark for rationalizing natural language analogical reasoning}.
\newblock In \emph{Findings of the Association for Computational Linguistics: ACL 2022}, pages 3941--3955, Dublin, Ireland. Association for Computational Linguistics.

\bibitem[{Chen et~al.(2024)Chen, Shuai, Zhang, Sun, and Cao}]{chen2024beyond}
Qing Chen, Wei Shuai, Jiyao Zhang, Zhida Sun, and Nan Cao. 2024.
\newblock Beyond numbers: Creating analogies to enhance data comprehension and communication with generative ai.
\newblock In \emph{Proceedings of the CHI Conference on Human Factors in Computing Systems}, pages 1--14.

\bibitem[{Cohen(1960)}]{cohen1960coefficient}
Jacob Cohen. 1960.
\newblock A coefficient of agreement for nominal scales.
\newblock \emph{Educational and psychological measurement}, 20(1):37--46.

\bibitem[{Conolly-Smith(2009)}]{7dd02d13-40ca-3017-bcff-8c8f4247ae25}
Peter Conolly-Smith. 2009.
\newblock \href {http://www.jstor.org/stable/40543352} {"connecting the dots": Munich, iraq, and the lessons of history}.
\newblock \emph{The History Teacher}, 43(1):31--51.

\bibitem[{Czinczoll et~al.(2022)Czinczoll, Yannakoudakis, Mishra, and Shutova}]{czinczoll-etal-2022-scientific}
Tamara Czinczoll, Helen Yannakoudakis, Pushkar Mishra, and Ekaterina Shutova. 2022.
\newblock \href {https://aclanthology.org/2022.findings-emnlp.153} {Scientific and creative analogies in pretrained language models}.
\newblock In \emph{Findings of the Association for Computational Linguistics: EMNLP 2022}, pages 2094--2100, Abu Dhabi, United Arab Emirates. Association for Computational Linguistics.

\bibitem[{Ding et~al.(2023)Ding, Srinivasan, MacNeil, and Chan}]{ding2023fluid}
Zijian Ding, Arvind Srinivasan, Stephen MacNeil, and Joel Chan. 2023.
\newblock Fluid transformers and creative analogies: Exploring large language models' capacity for augmenting cross-domain analogical creativity.
\newblock \emph{arXiv preprint arXiv:2302.12832}.

\bibitem[{Dobney(1974)}]{dobney1974lessons}
Fredrick~J Dobney. 1974.
\newblock " lessons" of the past: The use and misuse of history in american foreign policy.

\bibitem[{Falkenhainer et~al.(1989)Falkenhainer, Forbus, and Gentner}]{falkenhainer1989structure}
Brian Falkenhainer, Kenneth~D Forbus, and Dedre Gentner. 1989.
\newblock The structure-mapping engine: Algorithm and examples.
\newblock \emph{Artificial intelligence}, 41(1):1--63.

\bibitem[{Fournier et~al.(2020)Fournier, Dupoux, and Dunbar}]{fournier-etal-2020-analogies}
Louis Fournier, Emmanuel Dupoux, and Ewan Dunbar. 2020.
\newblock \href {https://doi.org/10.18653/v1/2020.conll-1.29} {Analogies minus analogy test: measuring regularities in word embeddings}.
\newblock In \emph{Proceedings of the 24th Conference on Computational Natural Language Learning}, pages 365--375, Online. Association for Computational Linguistics.

\bibitem[{Gao et~al.(2023)Gao, Xiong, Gao, Jia, Pan, Bi, Dai, Sun, and Wang}]{gao2023retrieval}
Yunfan Gao, Yun Xiong, Xinyu Gao, Kangxiang Jia, Jinliu Pan, Yuxi Bi, Yi~Dai, Jiawei Sun, and Haofen Wang. 2023.
\newblock Retrieval-augmented generation for large language models: A survey.
\newblock \emph{arXiv preprint arXiv:2312.10997}.

\bibitem[{Gentner(1983)}]{gentner1983structure}
Dedre Gentner. 1983.
\newblock Structure-mapping: A theoretical framework for analogy.
\newblock \emph{Cognitive science}, 7(2):155--170.

\bibitem[{Ghilani et~al.(2017)Ghilani, Luminet, Erb, Flassbeck, Rosoux, Tames, and Klein}]{ghilani2017looking}
Djouaria Ghilani, Olivier Luminet, Hans-Peter Erb, Christine Flassbeck, Val{\'e}rie Rosoux, Ismee Tames, and Olivier Klein. 2017.
\newblock Looking forward to the past: An interdisciplinary discussion on the use of historical analogies and their effects.
\newblock \emph{Memory Studies}, 10(3):274--285.

\bibitem[{Gladkova et~al.(2016)Gladkova, Drozd, and Matsuoka}]{gladkova-etal-2016-analogy}
Anna Gladkova, Aleksandr Drozd, and Satoshi Matsuoka. 2016.
\newblock \href {https://doi.org/10.18653/v1/N16-2002} {Analogy-based detection of morphological and semantic relations with word embeddings: what works and what doesn{'}t.}
\newblock In \emph{Proceedings of the {NAACL} Student Research Workshop}, pages 8--15, San Diego, California. Association for Computational Linguistics.

\bibitem[{Guldi and Armitage(2014)}]{guldi2014history}
Jo~Guldi and David Armitage. 2014.
\newblock \emph{The history manifesto}.
\newblock Cambridge University Press.

\bibitem[{Holyoak and Thagard(1996)}]{holyoak1996mental}
Keith~J Holyoak and Paul Thagard. 1996.
\newblock \emph{Mental leaps: Analogy in creative thought}.
\newblock MIT press.

\bibitem[{Jiayang et~al.(2023)Jiayang, Qiu, Chan, Fang, Wang, Chan, Ru, Guo, Zhang, Song, Zhang, and Zhang}]{jiayang-etal-2023-storyanalogy}
Cheng Jiayang, Lin Qiu, Tsz Chan, Tianqing Fang, Weiqi Wang, Chunkit Chan, Dongyu Ru, Qipeng Guo, Hongming Zhang, Yangqiu Song, Yue Zhang, and Zheng Zhang. 2023.
\newblock \href {https://doi.org/10.18653/v1/2023.emnlp-main.706} {{S}tory{A}nalogy: Deriving story-level analogies from large language models to unlock analogical understanding}.
\newblock In \emph{Proceedings of the 2023 Conference on Empirical Methods in Natural Language Processing}, pages 11518--11537, Singapore. Association for Computational Linguistics.

\bibitem[{Ju et~al.(2024)Ju, Sun, Du, Yuan, Ren, and Liu}]{ju2024large}
Tianjie Ju, Weiwei Sun, Wei Du, Xinwei Yuan, Zhaochun Ren, and Gongshen Liu. 2024.
\newblock How large language models encode context knowledge? a layer-wise probing study.
\newblock \emph{arXiv preprint arXiv:2402.16061}.

\bibitem[{Keulen(2023)}]{HistoricalAnalogies}
Sjoerd Keulen. 2023.
\newblock \href {https://doi.org/10.1163/25895893-bja10036} {Historical analogies: Functions, limitations and the correct use of historical analogies in applied history}.
\newblock \emph{Journal of Applied History}, 5(2):111 -- 131.

\bibitem[{Khong(2020)}]{khong2020analogies}
Yuen~Foong Khong. 2020.
\newblock \emph{Analogies at War: Korea, Munich, Dien Bien Phu, and the Vietnam Decisions of 1965}.
\newblock Princeton University Press.

\bibitem[{Kirchenbauer and Barns(2024)}]{kirchenbauer2024hallucination}
Jason Kirchenbauer and Caleb Barns. 2024.
\newblock Hallucination reduction in large language models with retrieval-augmented generation using wikipedia knowledge.

\bibitem[{Mikolov et~al.(2013)Mikolov, Chen, Corrado, and Dean}]{mikolov2013efficient}
Tomas Mikolov, Kai Chen, Greg Corrado, and Jeffrey Dean. 2013.
\newblock Efficient estimation of word representations in vector space.
\newblock \emph{arXiv preprint arXiv:1301.3781}.

\bibitem[{Neelakantan et~al.(2022)Neelakantan, Xu, Puri, Radford, Han, Tworek, Yuan, Tezak, Kim, Hallacy et~al.}]{neelakantan2022text}
Arvind Neelakantan, Tao Xu, Raul Puri, Alec Radford, Jesse~Michael Han, Jerry Tworek, Qiming Yuan, Nikolas Tezak, Jong~Wook Kim, Chris Hallacy, et~al. 2022.
\newblock Text and code embeddings by contrastive pre-training.
\newblock \emph{arXiv preprint arXiv:2201.10005}.

\bibitem[{Niwattanakul et~al.(2013)Niwattanakul, Singthongchai, Naenudorn, and Wanapu}]{niwattanakul2013using}
Suphakit Niwattanakul, Jatsada Singthongchai, Ekkachai Naenudorn, and Supachanun Wanapu. 2013.
\newblock Using of jaccard coefficient for keywords similarity.
\newblock In \emph{Proceedings of the international multiconference of engineers and computer scientists}, volume~1, pages 380--384.

\bibitem[{OpenAI(2022)}]{openai2022chatgpt}
OpenAI. 2022.
\newblock \href {https://openai.com/blog/chatgpt} {Chatgpt}.

\bibitem[{OpenAI(2023)}]{openai2023gpt4}
OpenAI. 2023.
\newblock \href {https://arxiv.org/abs/2303.08774} {Gpt-4 technical report}.
\newblock \emph{Preprint}, arXiv:2303.08774.

\bibitem[{Parsons and Nalau(2016)}]{parsons2016historical}
Meg Parsons and Johanna Nalau. 2016.
\newblock Historical analogies as tools in understanding transformation.
\newblock \emph{Global Environmental Change}, 38:82--96.

\bibitem[{Pearson(1920)}]{pearson1920notes}
Karl Pearson. 1920.
\newblock Notes on the history of correlation.
\newblock \emph{Biometrika}, 13(1):25--45.

\bibitem[{Petroni et~al.(2019)Petroni, Rockt{\"a}schel, Riedel, Lewis, Bakhtin, Wu, and Miller}]{petroni-etal-2019-language}
Fabio Petroni, Tim Rockt{\"a}schel, Sebastian Riedel, Patrick Lewis, Anton Bakhtin, Yuxiang Wu, and Alexander Miller. 2019.
\newblock \href {https://doi.org/10.18653/v1/D19-1250} {Language models as knowledge bases?}
\newblock In \emph{Proceedings of the 2019 Conference on Empirical Methods in Natural Language Processing and the 9th International Joint Conference on Natural Language Processing (EMNLP-IJCNLP)}, pages 2463--2473, Hong Kong, China. Association for Computational Linguistics.

\bibitem[{Rawte et~al.(2023)Rawte, Sheth, and Das}]{rawte2023survey}
Vipula Rawte, Amit Sheth, and Amitava Das. 2023.
\newblock A survey of hallucination in large foundation models.
\newblock \emph{arXiv preprint arXiv:2309.05922}.

\bibitem[{Renze and Guven(2024)}]{renze2024self}
Matthew Renze and Erhan Guven. 2024.
\newblock Self-reflection in llm agents: Effects on problem-solving performance.
\newblock \emph{arXiv preprint arXiv:2405.06682}.

\bibitem[{Schuman and Rieger(1992)}]{schuman1992historical}
Howard Schuman and Cheryl Rieger. 1992.
\newblock Historical analogies, generational effects, and attitudes toward war.
\newblock \emph{American Sociological Review}, pages 315--326.

\bibitem[{Shinn et~al.(2023)Shinn, Cassano, Gopinath, Narasimhan, and Yao}]{shinn2023reflexion}
Noah Shinn, Federico Cassano, Ashwin Gopinath, Karthik~R Narasimhan, and Shunyu Yao. 2023.
\newblock \href {https://openreview.net/forum?id=vAElhFcKW6} {Reflexion: language agents with verbal reinforcement learning}.
\newblock In \emph{Thirty-seventh Conference on Neural Information Processing Systems}.

\bibitem[{Shuster et~al.(2021)Shuster, Poff, Chen, Kiela, and Weston}]{shuster2021retrieval}
Kurt Shuster, Spencer Poff, Moya Chen, Douwe Kiela, and Jason Weston. 2021.
\newblock Retrieval augmentation reduces hallucination in conversation.
\newblock \emph{arXiv preprint arXiv:2104.07567}.

\bibitem[{Spearman(1961)}]{spearman1961proof}
Charles Spearman. 1961.
\newblock The proof and measurement of association between two things.

\bibitem[{Sultan et~al.(2024)Sultan, Bitton, Yosef, and Shahaf}]{sultan2024parallelparc}
Oren Sultan, Yonatan Bitton, Ron Yosef, and Dafna Shahaf. 2024.
\newblock Parallelparc: A scalable pipeline for generating natural-language analogies.
\newblock \emph{arXiv preprint arXiv:2403.01139}.

\bibitem[{Sultan and Shahaf(2022)}]{sultan-shahaf-2022-life}
Oren Sultan and Dafna Shahaf. 2022.
\newblock \href {https://aclanthology.org/2022.emnlp-main.232} {Life is a circus and we are the clowns: Automatically finding analogies between situations and processes}.
\newblock In \emph{Proceedings of the 2022 Conference on Empirical Methods in Natural Language Processing}, pages 3547--3562, Abu Dhabi, United Arab Emirates. Association for Computational Linguistics.

\bibitem[{Sung et~al.(2021)Sung, Lee, Yi, Jeon, Kim, and Kang}]{sung-etal-2021-language}
Mujeen Sung, Jinhyuk Lee, Sean Yi, Minji Jeon, Sungdong Kim, and Jaewoo Kang. 2021.
\newblock \href {https://doi.org/10.18653/v1/2021.emnlp-main.388} {Can language models be biomedical knowledge bases?}
\newblock In \emph{Proceedings of the 2021 Conference on Empirical Methods in Natural Language Processing}, pages 4723--4734, Online and Punta Cana, Dominican Republic. Association for Computational Linguistics.

\bibitem[{Tibshirani and Efron(1993)}]{tibshirani1993introduction}
Robert~J Tibshirani and Bradley Efron. 1993.
\newblock An introduction to the bootstrap.
\newblock \emph{Monographs on statistics and applied probability}, 57(1):1--436.

\bibitem[{Tonmoy et~al.(2024)Tonmoy, Zaman, Jain, Rani, Rawte, Chadha, and Das}]{tonmoy2024comprehensive}
SM~Tonmoy, SM~Zaman, Vinija Jain, Anku Rani, Vipula Rawte, Aman Chadha, and Amitava Das. 2024.
\newblock A comprehensive survey of hallucination mitigation techniques in large language models.
\newblock \emph{arXiv preprint arXiv:2401.01313}.

\bibitem[{Touvron et~al.(2023)Touvron, Lavril, Izacard, Martinet, Lachaux, Lacroix, Rozi{\`e}re, Goyal, Hambro, Azhar et~al.}]{touvron2023llama}
Hugo Touvron, Thibaut Lavril, Gautier Izacard, Xavier Martinet, Marie-Anne Lachaux, Timoth{\'e}e Lacroix, Baptiste Rozi{\`e}re, Naman Goyal, Eric Hambro, Faisal Azhar, et~al. 2023.
\newblock Llama: Open and efficient foundation language models.
\newblock \emph{arXiv preprint arXiv:2302.13971}.

\bibitem[{Turney and Littman(2005)}]{turney2005corpus}
Peter~D Turney and Michael~L Littman. 2005.
\newblock Corpus-based learning of analogies and semantic relations.
\newblock \emph{Machine Learning}, 60:251--278.

\bibitem[{Ushio et~al.(2021)Ushio, Espinosa~Anke, Schockaert, and Camacho-Collados}]{ushio-etal-2021-bert}
Asahi Ushio, Luis Espinosa~Anke, Steven Schockaert, and Jose Camacho-Collados. 2021.
\newblock \href {https://doi.org/10.18653/v1/2021.acl-long.280} {{BERT} is to {NLP} what {A}lex{N}et is to {CV}: Can pre-trained language models identify analogies?}
\newblock In \emph{Proceedings of the 59th Annual Meeting of the Association for Computational Linguistics and the 11th International Joint Conference on Natural Language Processing (Volume 1: Long Papers)}, pages 3609--3624, Online. Association for Computational Linguistics.

\bibitem[{Vrande{\v{c}}i{\'c} and Kr{\"o}tzsch(2014)}]{vrandevcic2014wikidata}
Denny Vrande{\v{c}}i{\'c} and Markus Kr{\"o}tzsch. 2014.
\newblock Wikidata: a free collaborative knowledgebase.
\newblock \emph{Communications of the ACM}, 57(10):78--85.

\bibitem[{Wang et~al.(2024)Wang, Wu, Wei, Jegelka, and Wang}]{wang2024theoretical}
Yifei Wang, Yuyang Wu, Zeming Wei, Stefanie Jegelka, and Yisen Wang. 2024.
\newblock A theoretical understanding of self-correction through in-context alignment.
\newblock \emph{arXiv preprint arXiv:2405.18634}.

\bibitem[{Webb et~al.(2022)Webb, Holyoak, and Lu}]{webb2022emergent}
Taylor Webb, Keith~J Holyoak, and Hongjing Lu. 2022.
\newblock Emergent analogical reasoning in large language models.
\newblock \emph{arXiv preprint arXiv:2212.09196}.

\bibitem[{West et~al.(2022)West, Bhagavatula, Hessel, Hwang, Jiang, Le~Bras, Lu, Welleck, and Choi}]{west-etal-2022-symbolic}
Peter West, Chandra Bhagavatula, Jack Hessel, Jena Hwang, Liwei Jiang, Ronan Le~Bras, Ximing Lu, Sean Welleck, and Yejin Choi. 2022.
\newblock \href {https://doi.org/10.18653/v1/2022.naacl-main.341} {Symbolic knowledge distillation: from general language models to commonsense models}.
\newblock In \emph{Proceedings of the 2022 Conference of the North American Chapter of the Association for Computational Linguistics: Human Language Technologies}, pages 4602--4625, Seattle, United States. Association for Computational Linguistics.

\bibitem[{Wijesiriwardene et~al.(2023)Wijesiriwardene, Wickramarachchi, Gajera, Gowaikar, Gupta, Chadha, Reganti, Sheth, and Das}]{wijesiriwardene-etal-2023-analogical}
Thilini Wijesiriwardene, Ruwan Wickramarachchi, Bimal Gajera, Shreeyash Gowaikar, Chandan Gupta, Aman Chadha, Aishwarya~Naresh Reganti, Amit Sheth, and Amitava Das. 2023.
\newblock \href {https://doi.org/10.18653/v1/2023.findings-acl.218} {{ANALOGICAL} - a novel benchmark for long text analogy evaluation in large language models}.
\newblock In \emph{Findings of the Association for Computational Linguistics: ACL 2023}, pages 3534--3549, Toronto, Canada. Association for Computational Linguistics.

\bibitem[{Xie et~al.(2024)Xie, Zhang, Chen, Lou, and Su}]{xie2024adaptive}
Jian Xie, Kai Zhang, Jiangjie Chen, Renze Lou, and Yu~Su. 2024.
\newblock \href {https://openreview.net/forum?id=auKAUJZMO6} {Adaptive chameleon or stubborn sloth: Revealing the behavior of large language models in knowledge conflicts}.
\newblock In \emph{The Twelfth International Conference on Learning Representations}.

\bibitem[{Xu et~al.(2024)Xu, Li, Tao, Shen, Cheng, Li, Xu, Tao, and Zhou}]{xu2024survey}
Xiaohan Xu, Ming Li, Chongyang Tao, Tao Shen, Reynold Cheng, Jinyang Li, Can Xu, Dacheng Tao, and Tianyi Zhou. 2024.
\newblock A survey on knowledge distillation of large language models.
\newblock \emph{arXiv preprint arXiv:2402.13116}.

\bibitem[{Ye et~al.(2024)Ye, Wang, Choi, Lu, Sharma, Shen, Tiyyala, Andrews, and Khashabi}]{ye2024analobench}
Xiao Ye, Andrew Wang, Jacob Choi, Yining Lu, Shreya Sharma, Lingfeng Shen, Vijay Tiyyala, Nicholas Andrews, and Daniel Khashabi. 2024.
\newblock Analobench: Benchmarking the identification of abstract and long-context analogies.
\newblock \emph{arXiv preprint arXiv:2402.12370}.

\bibitem[{Yuan et~al.(2023{\natexlab{a}})Yuan, Chen, Fu, Ge, Shah, Jankowski, Xiao, and Yang}]{yuan-etal-2023-distilling}
Siyu Yuan, Jiangjie Chen, Ziquan Fu, Xuyang Ge, Soham Shah, Charles Jankowski, Yanghua Xiao, and Deqing Yang. 2023{\natexlab{a}}.
\newblock \href {https://doi.org/10.18653/v1/2023.acl-long.236} {Distilling script knowledge from large language models for constrained language planning}.
\newblock In \emph{Proceedings of the 61st Annual Meeting of the Association for Computational Linguistics (Volume 1: Long Papers)}, pages 4303--4325, Toronto, Canada. Association for Computational Linguistics.

\bibitem[{Yuan et~al.(2023{\natexlab{b}})Yuan, Chen, Ge, Xiao, and Yang}]{yuan-etal-2023-beneath}
Siyu Yuan, Jiangjie Chen, Xuyang Ge, Yanghua Xiao, and Deqing Yang. 2023{\natexlab{b}}.
\newblock \href {https://doi.org/10.18653/v1/2023.findings-emnlp.160} {Beneath surface similarity: Large language models make reasonable scientific analogies after structure abduction}.
\newblock In \emph{Findings of the Association for Computational Linguistics: EMNLP 2023}, pages 2446--2460, Singapore. Association for Computational Linguistics.

\bibitem[{Yuan et~al.(2023{\natexlab{c}})Yuan, Chen, Sun, Liang, Xiao, and Yang}]{yuan2023analogykb}
Siyu Yuan, Jiangjie Chen, Changzhi Sun, Jiaqing Liang, Yanghua Xiao, and Deqing Yang. 2023{\natexlab{c}}.
\newblock Analogykb: Unlocking analogical reasoning of language models with a million-scale knowledge base.
\newblock \emph{arXiv preprint arXiv:2305.05994}.

\bibitem[{Yuan et~al.(2024)Yuan, Jiayang, Qiu, and Yang}]{yuan2024boosting}
Siyu Yuan, Cheng Jiayang, Lin Qiu, and Deqing Yang. 2024.
\newblock Boosting scientific concepts understanding: Can analogy from teacher models empower student models?
\newblock \emph{arXiv preprint arXiv:2406.11375}.

\bibitem[{Zhang et~al.(2023)Zhang, Li, Cui, Cai, Liu, Fu, Huang, Zhao, Zhang, Chen et~al.}]{zhang2023siren}
Yue Zhang, Yafu Li, Leyang Cui, Deng Cai, Lemao Liu, Tingchen Fu, Xinting Huang, Enbo Zhao, Yu~Zhang, Yulong Chen, et~al. 2023.
\newblock Siren's song in the ai ocean: a survey on hallucination in large language models.
\newblock \emph{arXiv preprint arXiv:2309.01219}.

\end{thebibliography}

\clearpage
\appendix
\begin{appendix}
\label{sec:appendix}


\section{Crowd-sourcing Details}\label{appendix:Crowd-sourcing}
\begin{figure*}[t]
    \centering
\includegraphics[width=\linewidth]{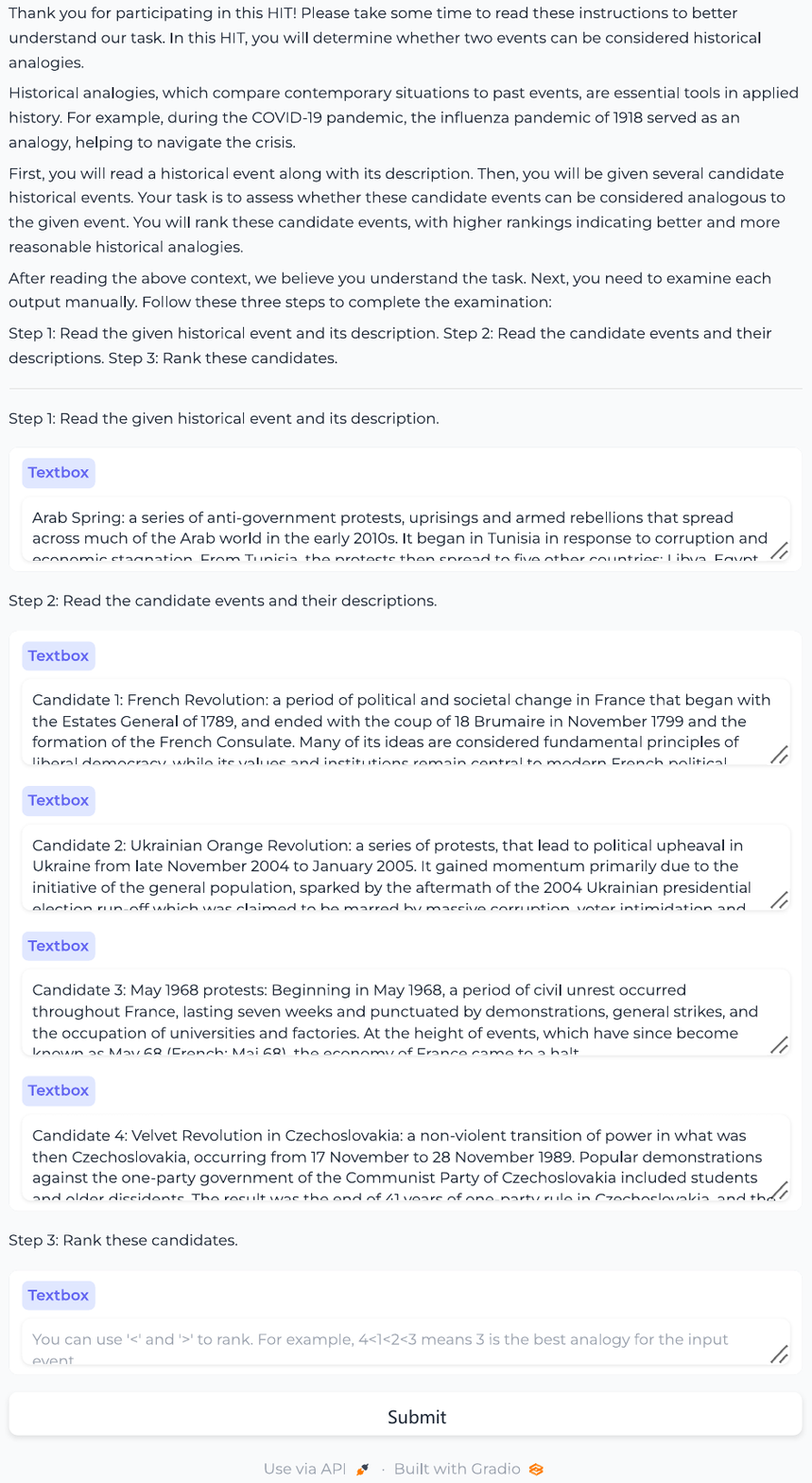}
    \caption{The screenshots of the instructions and interface for historical analogy manual annotation.}
    \label{fig:history_annotation}
\end{figure*}

We have recruited a team of three undergraduates from the history department.
To resolve conflicting annotations, we adopt a voting majority principle to determine the results.
Each annotator is compensated at \$8 per hour, which surpasses the local minimum wage.  
Screenshots of the instructions and interface for the annotation of historical analogies are shown in Figure~\ref{fig:history_annotation}.

\section{Resource of \method}\label{sec:resource}
We manually collect samples of popular analogies from web pages and papers related to historical analogies:
\begin{itemize}
    \item \url{https://psyche.co/guides/how-should-you-interpret-historical-analogies-in-the-popular-press}
    \item \url{https://scholars-stage.org/sino-american-competition-and-the-search-for-historical-analogies}
    \item \url{https://www.insidehighered.com/opinion/blogs/higher-ed-gamma/2024/03/08/use-and-misuse-historical-analogies}
    \item \url{https://www.sciencedirect.com/science/article/pii/S1090944316301636}
    \item \url{https://medium.com/@ella.ayalon/on-historical-analogies-3f253e52bfbc}
    \item \url{https://brill.com/view/journals/joah/5/2/article-p111_2.xml}
    \item \url{https://journals.sagepub.com/doi/abs/10.1177/1750698017701609}
    \item \url{https://aeon.co/essays/what-thucydides-really-thought-about-historical-analogies}
    \item \url{https://origins.osu.edu/history-news/historical-analogies-handle-care?}
    \item \url{https://www.emerald.com/insight/content/doi/10.1108/SSRP-03-2019-0020}
    \item \url{https://onlinelibrary.wiley.com/doi/epdf/10.1111/0162-895X.00145?saml_referrer}
    \item \url{https://www.washingtonpost.com/outlook/2021/05/03/historical-analogies-covid-fascism-mccarthyism/}
    \item \url{https://slate.com/news-and-politics/2014/07/pick-your-analogy-is-the-middle-east-today-more-like-world-war-i-the-cold-war-or-the-thirty-years-war.html}
    \item \url{https://launiusr.wordpress.com/2012/05/01/the-use-and-abuse-of-historical-analogs/}
    \item \url{https://www.usni.org/magazines/proceedings/2022/may/new-analogy-america-and-china-through-lens-pax-britannica}
\end{itemize}

\section{Prompt Template}\label{appendix:prompt}

\subsection{Prompt Template for Evaluation}\label{appendix:prompt_eval}
The prompt template for GPT-4 to summarize the dimensions of the event is shown in List~\ref{lst:instruction_prompt}


\lstset{
    backgroundcolor=\color[RGB]{245,245,244},
    breaklines=true,
    breakindent=0pt,
    basicstyle=\ttfamily\small,
    emph={Task,prompt,Examples,Test,Data},
    emphstyle={\bfseries\color{NavyBlue}}
}\begin{lstlisting}[caption={Instruction templates for GPT-4 to summarize the dimensions of the event.},label=lst:instruction_prompt]
/* Task prompt */
You are an event summary robot. For the event description input, please combine your knowledge and summarize it into four parts: topic, background, process and result. The summary should be concise, with each part consisting of only one sentence and no more than 100 words.
/* Examples */
Input Event: September 11 attacks
{Description of September 11 attacks}
Output: {Topic, Background, Process and Result of September 11 attacks}
/* Test Data */
Input Event: COVID-19 pandemic
{Description of COVID-19 pandemic}
Output: Topic: global health crises caused by viruses, resulting in widespread illness and significant mortality...
\end{lstlisting}



The prompt template for abstract similarity scoring is shown in List~\ref{lst:instruction_similarity}.


\lstset{
    backgroundcolor=\color[RGB]{245,245,244},
    breaklines=true,
    breakindent=0pt,
    basicstyle=\ttfamily\small,
    emph={Task,prompt,Examples,Test,Data,Evaluation,Criteria},
    emphstyle={\bfseries\color{NavyBlue}}
}\begin{lstlisting}[caption={Instruction templates for GPT-4 to score the abstract similarity for the given two events.},label=lst:instruction_similarity]
/* Task prompt */
You are a sentence-level analogy-scoring robot. Given the two descriptions, please judge the quality of the analogy and give it a score (1-4). The quality of an analogy only focuses on the abstract-level similarity, rather than the literal
similarity.
/* Evaluation Criteria */
1 point: The two descriptions belong to completely different
topics or fields, have no connection, and cannot be compared.
2 points: The two descriptions belong to the same general theme, but the specific situation or aspect they express is significantly different.
3 points: The two descriptions belong to the same topic and
express similar general situations, but differ somewhat in
details or focus.
4 points: The two descriptions pertain to the same topic, with
the general situation expressed being highly similar, and the
concepts and key points closely overlapping.
/* Test Data */
{Description of COVID-19 pandemic}
{Description of Spanish pandemic}
Score: 3
\end{lstlisting}




\subsection{Prompt Template for Methods}\label{appendix:prompt_method}
The prompt template of each method is given
in List~\ref{lst:method}
\newcolumntype{a}{>{\columncolor{BlueGreen!10}\centering\arraybackslash}p{2.1cm}}
\newcolumntype{b}{>{\columncolor{Blue!10}\centering\arraybackslash}p{2.1cm}}
\newcolumntype{d}{>{\columncolor{Purple!10}\centering\arraybackslash}p{1.8cm}}
\newcolumntype{q}{>{\columncolor{Gray!10}\centering\arraybackslash}p{1.8cm}}
\newcolumntype{e}{>{\columncolor{Brown!10}\centering\arraybackslash}p{1.8cm}}

\setlength\tabcolsep{1.3pt}
\begin{table*}[t]
  \centering
  \caption{Confidence intervals of experimental results on General Analogies, including Topic and Background dimensions.
  }
    \small
    \begin{tabular}{laaabbb}
    \toprule
    \multicolumn{1}{c}{\textbf{Method}} & \textbf{Topic$_{\texttt{Abs}}$} & \textbf{Topic$_{\texttt{Lit}}$} & \textbf{Topic$_{\texttt{All}}$}  & \textbf{Background$_{\texttt{Abs}}$} & \textbf{Background$_{\texttt{Lit}}$} & \textbf{Background$_{\texttt{All}}$} \\
    \midrule
    \multicolumn{7}{c}{GPT-3.5-Turbo} \\
    \midrule
    Direct Re. & {[}3.16, 3.43{]} & {[}0.17, 0.20{]} & {[}0.48, 0.57{]} & {[}2.86, 3.14{]} & {[}0.11, 0.15{]} & {[}0.63, 0.71{]} \\
    Two-stage Re. & {[}2.78, 3.09{]} & {[}0.16, 0.21{]} & {[}0.47, 0.56{]} & {[}2.54, 2.86{]} & {[}0.12, 0.17{]} & {[}0.54, 0.63{]} \\
    \cdashlinelr{1-7}
    Direct Gen. & {[}2.74, 3.04{]} & {[}0.12, 0.14{]} & {[}0.59, 0.66{]} & {[}2.54, 2.83{]} & {[}0.09, 0.11{]} & {[}0.61, 0.69{]} \\
    Two-stage Gen. & {[}3.07, 3.34{]} & {[}0.17, 0.23{]} & {[}0.52, 0.62{]} & {[}2.68, 2.97{]} & {[}0.13, 0.19{]} & {[}0.58, 0.67{]} \\
    Summarizing & {[}3.39, 3.61{]} & {[}0.16, 0.21{]} & {[}0.59, 0.68{]} & {[}2.89, 3.16{]} & {[}0.11, 0.16{]} & {[}0.64, 0.72{]} \\
    Self-reflection & {[}3.42, 3.62{]} & {[}0.16, 0.19{]} & {[}0.57, 0.65{]} & {[}3.10, 3.33{]} & {[}0.11, 0.13{]} & {[}0.69, 0.77{]} \\
    \midrule
    \multicolumn{7}{c}{Llama3.1-8B} \\
    \midrule
    Direct Re. & {[}3.16, 3.43{]} & {[}0.17, 0.20{]} & {[}0.48, 0.57{]} & {[}2.86, 3.14{]} & {[}0.11, 0.15{]} & {[}0.63, 0.71{]} \\
    Two-stage Re. & {[}2.97, 3.26{]} & {[}0.14, 0.17{]} & {[}0.61, 0.68{]} & {[}2.62, 2.94{]} & {[}0.10, 0.12{]} & {[}0.67, 0.74{]} \\
    \cdashlinelr{1-7}
    Direct Gen. & {[}3.32, 3.56{]} & {[}0.17, 0.21{]} & {[}0.55, 0.65{]} & {[}2.94, 3.21{]} & {[}0.12, 0.17{]} & {[}0.63, 0.72{]} \\
    Two-stage Gen. & {[}3.07, 3.36{]} & {[}0.13, 0.16{]} & {[}0.58, 0.67{]} & {[}2.77, 3.05{]} & {[}0.10, 0.12{]} & {[}0.65, 0.73{]} \\
    Summarizing & {[}3.27, 3.54{]} & {[}0.15, 0.18{]} & {[}0.56, 0.65{]} & {[}2.98, 3.25{]} & {[}0.10, 0.12{]} & {[}0.68, 0.75{]} \\
    Self-reflection & {[}3.36, 3.57{]} & {[}0.16, 0.20{]} & {[}0.58, 0.67{]} & {[}2.95, 3.22{]} & {[}0.11, 0.15{]} & {[}0.66, 0.74{]} \\
    \bottomrule
    \end{tabular}%
  \label{tab:condifence-interval-1}%
\end{table*}%

\setlength\tabcolsep{1.3pt}
\begin{table*}[t]
  \centering
  \caption{Confidence intervals of experimental results on General Analogies, including Process, Result, and the total score of the multi-dimensional similarity metric.
  }
    \small
    \begin{tabular}{ldddqqqe}
    \toprule
    \multicolumn{1}{c}{\textbf{Method}} & \textbf{Process$_{\texttt{Abs}}$} & \textbf{Process$_{\texttt{Lit}}$} & \textbf{Process$_{\texttt{All}}$}  & \textbf{Result$_{\texttt{Abs}}$} & \textbf{Result$_{\texttt{Lit}}$} & \textbf{Result$_{\texttt{All}}$} & \textbf{MDS} \\
    \midrule
    \multicolumn{8}{c}{GPT-3.5-Turbo} \\
    \midrule
    Direct Re. & {[}2.84, 3.11{]} & {[}0.10, 0.12{]} & {[}0.66, 0.74{]} & {[}2.86, 3.15{]} & {[}0.11, 0.14{]} & {[}0.62, 0.70{]} & {[}3.50, 3.78{]} \\
    Two-stage Re. & {[}2.46, 2.81{]} & {[}0.10, 0.14{]} & {[}0.55, 0.64{]} & {[}2.59, 2.92{]} & {[}0.12, 0.16{]} & {[}0.54, 0.63{]} & {[}3.01, 3.39{]} \\
    \cdashlinelr{1-8}
    Direct Gen. & {[}2.49, 2.76{]} & {[}0.08, 0.10{]} & {[}0.65, 0.72{]} & {[}2.66, 2.93{]} & {[}0.09, 0.11{]} & {[}0.64, 0.71{]} & {[}3.54, 3.84{]} \\
    Two-stage Gen. & {[}2.87, 3.15{]} & {[}0.11, 0.15{]} & {[}0.65, 0.74{]} & {[}2.87, 3.13{]} & {[}0.12, 0.16{]} & {[}0.63, 0.70{]} & {[}3.45, 3.82{]} \\
    Summarizing & {[}2.98, 3.24{]} & {[}0.10, 0.14{]} & {[}0.70, 0.78{]} & {[}2.94, 3.21{]} & {[}0.12, 0.15{]} & {[}0.63, 0.71{]} & {[}3.65, 4.00{]} \\
    Self-reflection & {[}3.04, 3.29{]} & {[}0.10, 0.12{]} & {[}0.71, 0.79{]} & {[}3.01, 3.26{]} & {[}0.11, 0.14{]} & {[}0.66, 0.73{]} & {[}3.79, 4.06{]} \\
    \midrule
    \multicolumn{8}{c}{Llama3.1-8B} \\
    \midrule
    Direct Re. & {[}2.84, 3.11{]} & {[}0.10, 0.12{]} & {[}0.66, 0.74{]} & {[}2.86, 3.15{]} & {[}0.11, 0.14{]} & {[}0.62, 0.70{]} & {[}3.50, 3.78{]} \\
    Two-stage Re. & {[}2.68, 2.96{]} & {[}0.09, 0.10{]} & {[}0.59, 0.66{]} & {[}2.59, 2.87{]} & {[}0.10, 0.12{]} & {[}0.54, 0.62{]} & {[}3.46, 3.75{]} \\
    \cdashlinelr{1-8}
    Direct Gen. & {[}2.91, 3.18{]} & {[}0.11, 0.15{]} & {[}0.68, 0.76{]} & {[}2.87, 3.14{]} & {[}0.12, 0.16{]} & {[}0.62, 0.70{]} & {[}3.54, 3.90{]} \\
    Two-stage Gen. & {[}2.78, 3.05{]} & {[}0.09, 0.11{]} & {[}0.69, 0.76{]} & {[}2.66, 2.94{]} & {[}0.10, 0.12{]} & {[}0.62, 0.69{]} & {[}3.61, 3.91{]} \\
    Summarizing & {[}2.88, 3.16{]} & {[}0.09, 0.11{]} & {[}0.71, 0.78{]} & {[}2.95, 3.23{]} & {[}0.11, 0.14{]} & {[}0.66, 0.73{]} & {[}3.75, 4.03{]} \\
    Self-reflection & {[}2.94, 3.21{]} & {[}0.10, 0.13{]} & {[}0.71, 0.79{]} & {[}2.99, 3.27{]} & {[}0.11, 0.14{]} & {[}0.66, 0.74{]} & {[}3.75, 4.07{]} \\
    \bottomrule
    \end{tabular}%
  \label{tab:condifence-interval-2}%
\end{table*}%


\lstset{
    backgroundcolor=\color[RGB]{245,245,244},
    breaklines=true,
    breakindent=0pt,
    basicstyle=\ttfamily\small,
    escapeinside={(*@}{@*)},
    morekeywords={Examples},
    keywordstyle={\bfseries\color{NavyBlue}}
}\begin{lstlisting}[caption={Instruction templates of different methods in historical analogy generation.},label=lst:method]
(*@\textbf{\color{NavyBlue}Direct Generation}@*):
You are a historical analogy bot. For input events, your goal is to find the most appropriate event to use for analogizing with the input.
/* Examples */
Input Event:
coronavirus pandemic: {Description of coronavirus pandemic}
Historical Analogies Events:
Spanish flu

(*@\textbf{\color{NavyBlue}Candidate Proposals in Two-Stage Method}@*):
You are a historical analogy candidate proposals robot. For input events, please consider the summary, background, process and results, output n historical events that are similar in many aspects above, and return them in list format.
/* Examples */
Input Event: 
coronavirus pandemic: {Description of coronavirus pandemic}
The 10 historical events that are similar with input:  
["Spanish flu pandemic","Asian flu pandemic","Hong Kong flu pandemic","AIDS pandemic","Ebola outbreak in West Africa","SARS outbreak","H1N1 influenza pandemic","MERS outbreak","Cholera pandemics","Plague pandemics"]

(*@\textbf{\color{NavyBlue}Selection in Two-Stage Method}@*):
You are an analogy robot. For the input event and the historical event used for selection, your goal is to find the best event that can be used for analogies. 
/* Examples */
Input Event:
coronavirus pandemic: {Description of coronavirus pandemic}
Optional Historical Events:
2022 South Asian floods: {Description of 2022 South Asian floods}
Croydon typhoid outbreak of 1937: {Description of Croydon typhoid outbreak}
Spanish flu: {Description of Spanish flu}
Cold War: {Description of Cold War}
Among the options, the most appropriate one to use as an analogy for coronavirus pandemic is Spanish flu

(*@\textbf{\color{NavyBlue}Candidate Proposals in Self-Reflection}@*):
You're a robot for proposing historical analogies events. Historical Analogy is comparsion of a known past event or person with a contemporary but unfamiliar event or person in order to identify common aspects between the two. For input events, please consider the summary, background, process and results, and output 5 historical events that are similar in many aspects above, and return them in list format. If there is any reflection, please modify the recommended events based on the reflection.
/* Examples */
Input Event: 
coronavirus pandemic: {Description of coronavirus pandemic}
The 10 historical events that are similar with input:  
["Spanish flu pandemic","Asian flu pandemic","Hong Kong flu pandemic","AIDS pandemic","Ebola outbreak in West Africa","SARS outbreak","H1N1 influenza pandemic","MERS outbreak","Cholera pandemics","Plague pandemics"]

(*@\textbf{\color{NavyBlue}Selection in Self-Reflection}@*):
You are a historical analogy reflection robot. Historical Analogy is comparsion of a known past event or person with a contemporary but unfamiliar event or person in order to identify common aspects between the two. For the input event and the candidate event set, please make a comparison, reflect on the shortcomings of the candidate set, and make suggestions for obtaining a better analogous candidate set. Suggestions should be succinct and concise, with a single sentence indicating the direction of change for the candidate set.
/* Examples */
===== Case 1
Input Event:
coronavirus pandemic: {Description of coronavirus pandemic}
Optional Historical Events:
2022 South Asian floods: {Description of 2022 South Asian floods}
Croydon typhoid outbreak of 1937: {Description of Croydon typhoid outbreak}
Thought:
The coronavirus pandemic is a global epidemic, so the themes of 2022 South Asian floods are completely different. The Croydon typhoid outbreak of 1937 was smaller in scope, while the coronavirus pandemic were global influenza pandemics, so there is no suitable analogy here and I need to reflect.        
Reflection:
Candidate events need to focus on the epidemic and its impact on a global scale.
===== Case 2
Input Event:
coronavirus pandemic: {Description of coronavirus pandemic}
Optional Historical Event:
Spanish flu: {Description of Spanish flu}
Cold War: {Description of Cold War}
Thought:
The Cold War has nothing to do with the epidemic. The Spanish flu is also an epidemic and has had a great impact in Europe, so it is a qualified analogy for the coronavirus pandemic.
Final Answer:
Spanish flu
\end{lstlisting}




\section{Stability Analysis of Experimental Results}\label{appendix:stability}
To demonstrate the stability and reliability of our experimental results, we use bootstrapping~\cite{tibshirani1993introduction} with 1000 resamples on general analogy to compute the 95\% confidence intervals for each metric.
Our results are shown in Tables~\ref{tab:condifence-interval-1} and Table~\ref{tab:condifence-interval-2}.
The confidence intervals align with the main experiment (Table~\ref{tab:main-result}) and show only about a 10\% fluctuation.
Therefore, our results are stable and trustworthy.
\end{appendix}

\end{document}